\documentclass[11pt,oneside]{article}

\usepackage[left=1in,right=1in,top=1in,bottom=1in]{geometry}
\usepackage{authblk}

\usepackage{setspace}
\usepackage[parfill]{parskip}
\usepackage{lscape}

\usepackage{color}
\usepackage[colorlinks,
            citecolor=blue,
            urlcolor=blue,
            linkcolor=blue]{hyperref}

\usepackage[utf8]{inputenc} 
\usepackage[T1]{fontenc}    
\usepackage{hyperref}       
\usepackage{url}            
\usepackage{booktabs}       
\usepackage{amsfonts}       
\usepackage{nicefrac}       
\usepackage{microtype}      
\usepackage{enumitem}
\usepackage{multirow}
\usepackage{algorithm}
\usepackage{nameref}
\usepackage{mathrsfs}
\usepackage{comment}
\usepackage{tablefootnote}
\usepackage[font=footnotesize]{caption}

\usepackage{microtype}
\usepackage{graphicx}
\usepackage{subfig}
\usepackage{amsmath}   
\usepackage{amsthm}
\usepackage{amssymb}
\usepackage{siunitx}   
\usepackage{bm}        
\usepackage{thmtools}  
\usepackage{thm-restate}
\usepackage{color}     
\usepackage{xspace}    

\newcommand{\ba}{\bm{a}}
\newcommand{\bb}{\bm{b}}
\newcommand{\bc}{\bm{c}}
\newcommand{\be}{\bm{e}}
\newcommand{\bu}{\bm{u}}
\newcommand{\bv}{\bm{v}}

\newcommand{\bx}{\bm{x}}

\newcommand{\bA}{\bm{A}}
\newcommand{\bB}{\bm{B}}

\newcommand{\bD}{\bm{D}}
\newcommand{\bE}{\bm{E}}
\newcommand{\balpha}{\bm{\alpha}}
\newcommand{\bSigma}{\bm{\Sigma}}
\newcommand{\bI}{\bm{I}}
\newcommand{\bJ}{\bm{J}}

\newcommand{\bO}{\bm{O}}
\newcommand{\bS}{\bm{S}}
\newcommand{\bU}{\bm{U}}
\newcommand{\bV}{\bm{V}}
\newcommand{\bX}{\bm{X}}

\newcommand{\bzero}{\bm{0}}
\newcommand{\Ccal}{\mathcal{C}}
\newcommand{\Ecal}{\mathcal{E}}

\newcommand{\Mcal}{\mathcal{M}}
\newcommand{\Ocal}{\mathcal{O}}
\newcommand{\Xcal}{\mathcal{X}}

\newcommand{\Vcal}{\mathcal{V}}
\newcommand{\Ycal}{\mathcal{Y}}

\newcommand{\R}{\mathbb{R}}
\newcommand{\B}{\mathbb{B}}

\newcommand{\bone}{\bm{1}}

\DeclareMathOperator*{\argmin}{arg\,min\,}
\DeclareMathOperator*{\argmax}{arg\,max\,}

\DeclareMathOperator{\Cov}{Cov}

\definecolor{ForestGreen}{RGB}{34,139,34}

\usepackage{pifont}
\newcommand{\cmark}{\ding{51}}%
\newcommand{\xmark}{\ding{55}}%

\DeclareMathOperator{\rank}{rank}
\DeclareMathOperator{\vecop}{vec}
\DeclareMathOperator{\Tr}{Tr\,}

\newcommand{\scotus}{\textsc{scotus}\xspace}
\newcommand{\cusum}{\textsc{cusum}\xspace}

\newcommand{\uvec}[1]{\overline{\vecop}\left\{#1\right\}}
\newcommand{\unuvec}[1]{\overline{\vecop}^{-1}\left\{#1\right\}}

\newtheorem{appendixtheorem}{Theorem}

\newtheorem{appendixproposition}{Proposition}

\newtheorem{appendixremark}{Remark}

\newcommand{\opnorm}[1]{\left\|#1\right\|_{\text{op}}}
\newcommand{\ropnorm}[1]{\left\|#1\right\|_{r\text{-op}}}
\newcommand{\ttm}[1]{\,\times_{#1}\,}
\newcommand{\ttv}[1]{\,\bar{\times}_{#1}\,}

\usepackage[natbib=true,
            maxcitenames=2,
            maxbibnames=20,
            citestyle=authoryear,
            bibstyle=authoryear,
            uniquename=false,
            uniquelist=false]{biblatex}
\renewbibmacro{in:}{}
\DeclareCiteCommand{\citeyearparlink}
    {}
    {\mkbibparens{\bibhyperref{\printdate}}}
    {\multicitedelim}
    {}
\newcommand{\citegenitive}[1]{\citeauthor{#1}'s \citeyearparlink{#1}}

\title{Multivariate Analysis for Multiple Network Data via Semi-Symmetric Tensor PCA}

\author[1]{Michael Weylandt}
\author[1,2]{George Michailidis}
\affil[1]{University of Florida Informatics Institute}
\affil[2]{Department of Statistics, University of Florida}

\usepackage{lineno}

\begin{document}

\maketitle
\begin{abstract}
Network data are commonly collected in a variety of applications, representing either directly measured or statistically inferred connections between features of interest. In an increasing number of domains, these networks are collected over time, such as interactions between users of a social media platform on different days, or across multiple subjects, such as in multi-subject studies of brain connectivity. When analyzing multiple large networks, dimensionality reduction techniques are often used to embed networks in a more tractable low-dimensional space. To this end, we develop a framework for principal components analysis (PCA) on collections of networks via a specialized tensor decomposition we term Semi-Symmetric Tensor PCA or SS-TPCA. We derive computationally efficient algorithms for computing our proposed SS-TPCA decomposition and establish statistical efficiency of our approach under a standard low-rank signal plus noise model. Remarkably, we show that SS-TPCA achieves the same estimation accuracy as classical matrix PCA, with error proportional to the square root of the number of vertices in the network and not the number of edges as might be expected. Our framework inherits many of the strengths of classical PCA and is suitable for a wide range of unsupervised learning tasks, including identifying principal networks, isolating meaningful changepoints or outlying observations, and for characterizing the ``variability network'' of the most varying edges. Finally, we demonstrate the effectiveness of our proposal on simulated data and on an example from empirical legal studies. The techniques used to establish our main consistency results are surprisingly straightforward and may find use in a variety of other network analysis problems.


\end{abstract}

\onehalfspace

\begin{refsection}

\section{Introduction} \label{sec:introduction}
Principal Components Analysis (PCA) is a fundamental tool for the analysis of multivariate data, enabling a wide range of dimension reduction, pattern recognition, and visualization strategies. Originally introduced by \citet{Pearson:1901} and \citet{Hotelling:1933} for investigating low-dimensional data in a Euclidean space, PCA has been generalized to much more general contexts, including data taking values in general inner product spaces \citep{Eaton:2007}, (possibly infinite-dimensional) function spaces \citep{Silverman:1996}, discrete and compositional spaces \citep{Liu:2018}, and shape spaces \citep{Jung:2012}. In this work, we analyze PCA-type decompositions for  \textit{network data}: that is, given a collection of undirected networks on a common set of nodes, we seek to identify common patterns in the form of one or more ``principal networks'' that capture most of the variation in our data. Our main technical tool is a \emph{tensor decomposition} on so-called \emph{semi-symmetric tensors}, for which we provide an efficient computational algorithm and rigorous statistical guarantees, in the form of a general finite sample consistency result. As we show below, our approach achieves the same convergence rates as classical (Euclidean) PCA up to a logarithmic factor, with error proportional to the \emph{square root} of the number of vertices rather than the number of edges, providing significant improvements when analyzing networks of even moderate size.

\begin{figure}[thb]
    \centering
    \includegraphics[width=\textwidth]{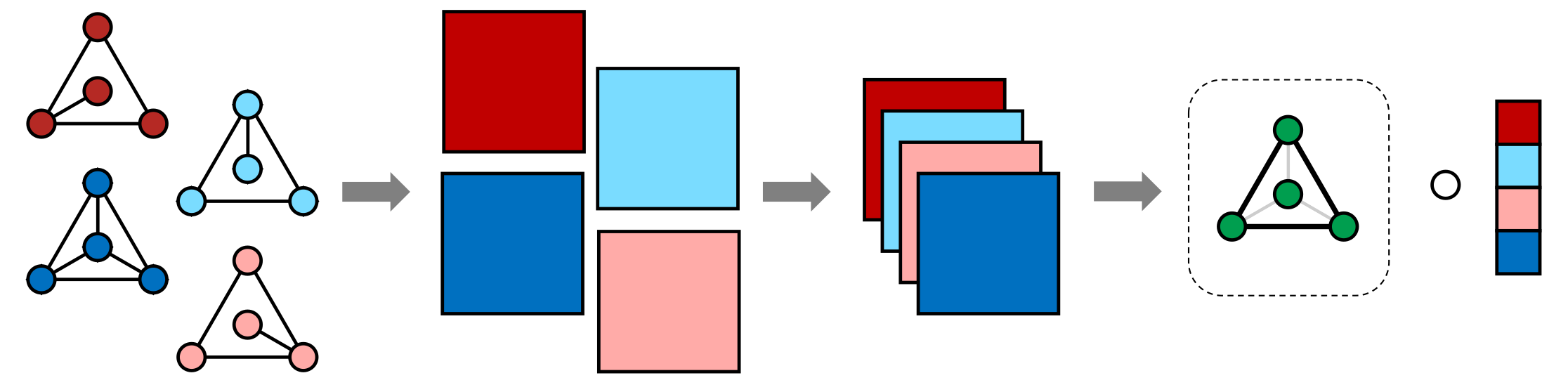}
    \caption{Analysis of Multiple Networks Using Semi-Symmetric Tensor PCA: Given a collection of networks on a common node set (left panel), we express each network using a common matrix representation, \emph{e.g.}, adjacency matrix or (regularized) graph Laplacian (second panel), which can then be formed into a semi-symmetric tensor(third panel). The Semi-Symmetric Tensor PCA decomposition (right panel) approximates this tensor as the product of two terms which can be interpreted as a ``principal network'' on the same node set (green network) and a loading vector across the observations.}
    \label{fig:schematic}
\end{figure}

Network-valued data are of increasing importance in analyzing modern structured data sets. In many situations, multiple networks are observed in conjunction with the same phenomenon, typically when independent replicates are observed over time (``network series'') or when different network representations of the same underlying dynamics can be observed \citep[][``multilayer networks'']{Kivela:2014}. Notable examples of this type of data can be found in neuroscience \citep{Zhang:2019}, group social dynamics \citep{Eagle:2009}, international development \citep{Hafner-Burton:2009}, and transportation studies \citep{Cardillo:2013}. Because these undirected (weighted) networks are observed on a common known set of nodes, they can each be represented as symmetric matrices and these symmetric matrices can be stacked into a third-order \emph{semi-symmetric} tensor, \emph{i.e.}, a tensor each slice of which is a symmetric matrix. In this paper, we study the decomposition depicted in Figure \ref{fig:schematic}: that is, decomposing a semi-symmetric tensor into a ``principal network'' term as well as a loading vector. More details of this decomposition follow in Section \ref{sec:sstpca}.

In this paper, we focus on scenarios where networks are directly observed and the structure of the networks is of primary interest. We distinguish this from closely related problems arising in the analysis of network-structured signals, \emph{i.e.}, graph signal processing \citep{Shuman:2013}, or in estimating network structure from observed multivariate data, \emph{i.e.}, structure learning of probabilistic graphical models \citep[Part III]{Maathuis:2018}. 

\subsection{Selected Background and Related Work}\label{subsec:related_work}
We take the ``canonical polyadic'' or ``CP'' tensor decomposition framework as our starting point. The CP decomposition approximates a tensor by a sum of rank-1 components, which are typically estimated using an alternating least squares approach \citep[Section 3]{Kolda:2009}. The CP decomposition has been applied as the basis for various forms of tensor PCA, where it is preferred over Tucker-type decompositions because it provides interpretable and ordered components, similar to those resulting from classical (matrix) PCA \citep{Zare:2018}. For scalar data arranged in a tensor, CP-based tensor PCA exhibits strong performance and has been fruitfully extended to sparse and high-dimensional settings with efficient algorithms \citep{Allen:2013,Allen:2012}, 
but it is less ideal for network settings. Specifically, CP-based approaches decompose a tensor into rank-1 components, but realistic networks essentially never have exact rank-1 structure, as this would imply a uniform connectivity pattern across all nodes. This problem---developing an efficient tensor decomposition and PCA framework which captures meaningful network structure---motivates the ``semi-symmetric tensor PCA'' (SS-TPCA) decomposition at the heart of this paper.

While computational theory for tensor decompositions is well-established, comprehensive \emph{statistical} theory remains an active area of research in the statistics and machine learning communities. \citet{Anandkumar:2014} give a general analysis of statistical estimation under CP-type models, developing incoherence conditions that allow recovery of multiple CP factors. Extending their analysis to the sparse CP framework of \citet{Allen:2012}, \citet{Sun:2016} establish consistency and sparse recovery under additional sparsity and incoherence assumptions. Consistency results have also been established for tensor decompositions using the Tucker \citep{Zhang:2018,Zhang:2019-JASA} and Tensor-Train \citep{Oseledets:2011,Zhou:2022} frameworks. A recent paper by \citet{Han:2022} considers general estimation problems under a low Tucker-rank assumption and establishes convergence guarantees for a wide range of problems, including low-rank approximation, tensor completion, and tensor regression. For tensors with symmetric structure, Anandkumar and co-authors \citep{Anandkumar:2014b,Janzamin:2019} analyze decompositions of fully-symmetric tensors, such as those arising in certain method-of-moments estimation schemes. The fully-symmetric case imposes additional computational concerns not present in the general case: \citet{Kolda:2015} discusses these issues and reviews related work in more detail. The tensor statistical estimation literature is too large to be comprehensively described here and we refer the reader to the recent survey by \citet{Bi:2021} for additional references.

While we are the first to provide rigorous statistical theory for semi-symmetric tensor decompositions, these techniques have already found wide use in the literature. In particular, our SS-TPCA builds upon the Network Tensor PCA proposal of \citet{Zhang:2019} who apply tensor techniques to brain connectivity in order to find connectomic structures correlated with a variety of interesting behavioral traits. \citet{Winter:2020} extend this approach to matrices of allow for graphs of different granularity, yielding an approach which is robust to the particular brain parcellation used. \citet{Wang:2014-CP_Journal} use a different semi-symmetric decomposition to perform Independent Components Analysis on spectroscopy data. Despite the empirical successes of these approaches, this work is the first to provide rigorous theoretical guarantees of the type presented in Theorem \ref{thm:consistency} below.

Separately, the task of finding common low-rank subspace representations of multiple graphs has been well-studied in the network science community. The recent work of \citet{Wang-PAMI:2021} is particularly closely related to our work: both papers consider a tensor decomposition approach to learning common features among a set of graphs. Our results significantly improve upon their findings, as we are able to give finite-sample results on the quality of both the estimated principal components and the loading vector under a  general ``low-rank'' model, while they only give an asymptotic consistency result for the principal component under a particular rank-1 variant of the Erd\H{o}s-R\'enyi model.  Similarly, the COSIE proposal of \citet{Arroyo:2021} develops a spectral embedding approach for multiple graphs, but does so leveraging techniques from distributed estimation rather than tensor decompositions and under a different probabilistic model than we consider below.

Our approach is motivated by the statistical analysis of multiple networks, an area of much recent activity. In addition to the embedding analysis described above, we highlight several recent advances in this space including: spectral clustering of network-valued data \citep{Mukherjee:2017}; model-based clustering of network valued data \citep{Matias:2017};
probabilistic models and Bayesian estimators for network series \citep{Sewell:2015,Gollini:2016,Durante:2017}; factor models for matrix time series \citep{Chen:2022}; two-sample testing \citep{Ghoshdastidar:2020}; and scalar-on-network regression \citep{Relion:2019,Guha:2021}. We expect that a robust Network PCA framework, such as that associated with our approach, can lead to advances in many of these objectives.

\subsection{Notation} \label{subsec:notation}

Notation used in this paper follows that of \citet{Kolda:2009} with two minor modifications: 
firstly, in the tensor-matrix product, $\ttm{k}$, we omit a transpose implicitly used by \citeauthor{Kolda:2009}: for two matrices $\bA, \bB$ we have $\bA \ttm{2} \bB = \bA\bB$ while they have $\bA \ttm{2} \bB = \bA\bB^T$. Consequently, we have $\Xcal \ttm{k} \bA \ttm{k} \bB = \Xcal \ttm{k} (\bA\bB)$ rather than $\Xcal \ttm{k} (\bB\bA)$. In general, the appropriate axes on which to multiply are clear from context. Secondly, while $\circ$ denotes the general outer product, we also adopt the convention that when it is applied to two matrices of equivalent size, it denotes a matrix product with its transpose, not a higher order product: that is, if $\bV \in \R^{p \times k}$, $\bV \circ \bV = \bV\bV^T$, a $p \times p$ matrix, not a $p \times k \times p \times k$ tensor. Unless otherwise noted, inner products and norms refer to the Frobenius norm and associated inner product ($\|\Xcal\| = \|\vecop(\Xcal)\|_{\ell_2}$ and $\langle \Xcal, \Ycal\rangle = \sum_i \vecop(\Xcal)_i\vecop(\Ycal)_i$).  We denote the (compact) Stiefel manifold by $\Vcal^{p \times r} = \{\bV \in \R^{p \times r}: \bV^T\bV = \bI_{r \times r}\}$. We will make frequent use of the fact that, for any $\bV \in \Vcal^{p \times r}$, $\bV\circ \bV$ is a rank-$r$ matrix of size $p \times p$.

Semi-symmetric tensors are third-order tensors, \emph{i.e.}, three dimensional arrays, each slice of which along a fixed axis yields a (real) symmetric matrix. By convention, we take the first two dimensions to be the axes of symmetry, so $\Xcal \in \R^{p \times p \times T}$ is a semi-symmetric tensor if $\Xcal_{\cdot\cdot i}$ is a $p \times p$ symmetric matrix for all $1 \leq i \leq T$. We introduce several notations for non-standard tensor operations which arise naturally in the semi-symmetric context: for a $p \times p \times T$ semi-symmetric tensor $\Xcal$ and a $p \times k$ matrix $\bV$, we define the trace-product $[\Xcal; \bV]$ as the $T$-vector whose $j\textsuperscript{th}$ element is given by $\langle \Xcal_{\cdot\cdot j}, \bV\bV^T\rangle = \Tr(\bV^T\Xcal_{\cdot\cdot j}\bV)$. For a $p\times p$ symmetric matrix $\bA$, $\uvec{\bA}$ denotes the $\binom{p}{2}$-vector formed from the upper triangle of $\bA$; $\unuvec{\cdot}$ denotes the inverse vector-to-symmetric matrix mapping. Finally, the operator norm $\opnorm{\cdot}$ of a symmetric matrix is given by the absolute value of the largest magnitude eigenvalue and the rank-$r$ operator norm of a semi-symmetric tensor is defined as $\ropnorm{\Xcal} = \max_{\bu, \bv \in \overline{\B}^T \times \Vcal^{p \times r}} \left|\left\langle [\Xcal; \bV], \bu\right\rangle\right|$, where $\overline{\B}^T$ is the unit ball in $\R^T$. In general, the semi-symmetric tensor operator norm is difficult to compute, but Proposition \ref{prop:ropnorm} in Section \ref{sec:theory} gives a tractable upper bound in terms of the operator norm of the individual slices of $\Xcal$.

\subsection{Our Contributions}\label{subsec:contributions}
Our contributions are threefold: firstly, building on a proposal of \citet{Zhang:2019}, we develop a computationally efficient flexible tensor PCA framework that is able to characterize networks arising from a large class of random graph models; secondly, we establish statistical consistency of the SS-TPCA procedure under fairly weak and general assumptions, centered on a tensor analogue of the low-rank mean model; finally, we establish a novel connection between multilayer network analysis and regularized $M$-estimation that is of significant independent theoretical interest.

The remainder of this paper is organized as follows: Section \ref{sec:sstpca} introduces the ``Semi-Symmetric Tensor PCA'' decomposition (SS-TPCA) and proposes an efficient algorithm for computing the SS-TPCA solution; in this section, we also discuss several useful aspects of the SS-TPCA framework, including procedures for computing multiple principal components and an extension to a functional PCA-type setting. Section \ref{sec:theory} further examines the theoretical properties of the SS-TPCA via a simple, yet novel, analytical framework we expect may find use in other contexts; Section \ref{sec:rank_pca} connects these results to the wider literature on regularized Principal Components Analysis and shows how the technical tools we develop can be used for other $M$-estimation problems arising from network-valued data. Section \ref{sec:empirical} demonstrates the usefulness and flexibility of the SS-TPCA framework, both in simulation studies and in an extended case study from empirical legal studies. Finally, Section \ref{sec:discussion} concludes the paper and discusses several possible future directions for investigation.


\section{Semi-Symmetric Tensor Principal Components Analysis} \label{sec:sstpca}

Our goal is to develop a tensor PCA framework which is able to capture realistic network structures in its output. While real-world networks have a variety of structures, we focus here on networks which are (approximately) low-rank, as these are most well-suited for PCA-type analyses. Many network models are approximately low-rank in expectation, most notably, the random dot-product graph family \citep{Athreya:2018}, which includes both latent position and stochastic block models as special cases. These models have found use in a wide range of contexts, including analysis of social media networks, brain connectivity networks, and genomic networks \citep{Sussman:2014,Lyzinski:2016}.

With these low-rank models in mind, we introduce our ``Semi-Symmetric Tensor PCA'' (SS-TPCA) factorization, which approximates a $p \times p \times T$ semi-symmetric tensor $\Xcal$ by a rank-$r$ ``principal component'' and a loading vector for the final mode, which we typically take to represent independent replicates observed over time. As seen in the sequel, it is natural in many applications to interpret these as a ``principal network'' and a ``time factor,'' but this viewpoint is not essential for the technical developments of this section.

The single-factor rank-$r$ SS-TPCA approximates $\Xcal$ as 
\[\Xcal \approx d\, \bV \circ \bV \circ \bu\]
where $\bu \in \R^T$ is a unit-norm $T$-vector, $d \in \R_{\geq 0}$ is a scale factor, and $\bV \in \Vcal^{p \times r}$ is an orthogonal matrix satisfying $\bV^T\bV = \bI_{r \times r}$. When $r = 1$ and $\bV_* = \bv_*$ is a unit vector, this coincides with the Network PCA approach of \citet{Zhang:2019}. This decomposition is related to the standard $r$-term CP decomposition: 
\[\Xcal \approx \sum_{i=1}^r \lambda_i \ba_i \circ \bb_i \circ \bc_i\] with the additional restrictions $\ba_i = \bb_i$ and $\bc_i = \bc_1$ for all $1 \leq i \leq r$ as well as $\langle \ba_i, \ba_j \rangle = \delta_{ij}$ to ensure orthogonality. These constraints restrict the number of free parameters from $r(2p + T + 1)$ to $pr + T - r(r + 1)/2 + 1$, which significantly improves both computational performance and estimation accuracy, as will be described in more detail below. 

A more general SS-TPCA, which we denote as the $(r_1, \cdots, r_K)$-SS-TPCA, may be defined by
\[\Xcal \approx \sum_{k=1}^K d_k \bV_k \circ \bV_k \circ \bu_k,\] where, as before, each $\bu_k$ is a unit-norm $T$-vector, $d_k$ are positive scale factors, and $\bV_k \in \Vcal^{p \times r_k}$ are orthogonal matrices used to construct the low-rank principal networks. We emphasize that the $\bV_k$ terms need not be of the same dimension, allowing our model to flexibly capture principal components of differing rank. The general SS-TPCA model inherits properties of both the popular CP and Tucker decompositions: it approximates $\Xcal$ by the sum of multiple simple components like the CP decomposition but allows for orthogonal low-rank factors like the Tucker decomposition.

To compute the single-factor SS-TPCA, we seek the best approximation of $\Xcal$ in the tensor Frobenius norm: 
\begin{align*}
    \argmin_{\bu, \bV, d}\text{   } &\|\Xcal - d\, \bV\circ \bV \circ \bu\|_F^2 \quad  
    \text{subject to }\bu \in \R^T, \|\bu\| = 1,
     d \in \R_{\geq 0},
     \bV \in \Vcal^{p \times r}
\end{align*}
It is easily seen that solving this problem is equivalent to minimizing the inner product: 
\[\langle \Xcal, \bV \circ\bV \circ \bu \rangle = \langle \Xcal \ttv{3} \bu, \bV\circ \bV\rangle\]
While this problem is difficult to optimize jointly in $\bu$ and $\bV$, closed-form global solutions are available both for $\bu$ with $\bV$ held constant and for $\bV$ with $\bu$ held constant. This motivates an alternating minimization (block coordinate descent) strategy comprised of alternating $\bu$ and $\bV$ updates. Specifically, holding $\bu$ constant, the optimal value of $\bV$ is simply the principal $r$ eigenvalues of $\Xcal \ttv{3} \bu$ and, holding $\bV$ constant, the optimal value of $\bu$ is a unit vector in the direction of $[\Xcal; \bV]$. Note that, for the single factor rank-1 SS-TPCA, the $\bV$-update simplifies to $\bu \propto \Xcal \ttv{1} \bv \ttv{2} \bv$. Putting these steps together, we obtain the following algorithm for the single-factor rank-$r$ SS-TPCA decomposition:
\begin{algorithm}[H]
\begin{itemize}
\item Input: $\Xcal$, $\bu^{(0)}$
\item Initialize: $k = 0$
\item Repeat until convergence:
\begin{enumerate}
\item[(i)] $\bV^{(k+1)} = \textsf{$r$-eigen}(\Xcal \ttv{3} \bu^{(k)})$
\item[(ii)] $\bu^{(k + 1)} = \textsf{Norm}([\Xcal; \bV^{(k+1)}])$
\item[(iii)] $k := k + 1$
\end{enumerate}
\item Return $\hat{\bu} = \bu^{(k)}$, $\hat{\bV} = \bV^{(k)}$, $\hat{d} = r^{-1}\langle \Xcal, \hat{\bV} \circ \hat{\bV} \circ \hat{\bu} \rangle_F$, and $\hat{\Xcal} = \hat{d}\, \hat{\bV} \circ \hat{\bV} \circ \hat{\bu}$.
\end{itemize}
\caption{Alternating Minimization Algorithm for the Rank-$r$ Single-Factor SS-TPCA}
\label{alg:single_factor_sstpca}
\end{algorithm}
\noindent where $\textsf{Norm}(\bx) = \bx / \|\bx\|_2$ and $\textsf{$r$-eigen}(\bX)$ denotes the first $r$ eigenvectors of $\bX$. When $\bX$ has both positive and negative eigenvalues, $\textsf{$r$-eigen}(\bX)$ denotes those eigenvectors whose eigenvalues have the largest magnitudes. Convergence of Algorithm \ref{alg:single_factor_sstpca} follows from the general block coordinate analysis of \citet{Tseng:2001} or from a minor extension of the results of \citet{Kofidis:2002}, noting that the use of an eigendecomposition allows the $\bV$-update to be solved to global optimality. In Section \ref{sec:rank_pca} below, we outline an alternate convergence proof based on recent developments in the theory of regularized (matrix) PCA.

We note here that we consider the case where $\bV \circ \bV$ is a projection matrix, \emph{i.e.} one with all eigenvalues either 0 or 1. This simplifies the derivation above and the theoretical analysis presented in the next section, but is not necessary for our approach. In situations where the eigenvalues of $\bV \circ \bV$ are allowed to vary, the $\bV$-update in Algorithm \ref{alg:single_factor_sstpca} can be replaced by $\bV^{(k+1)} = \bD^{1/2} \textsf{$r$-eigen}(\Xcal \ttv{3} \bu^{(k)})$ where $\bD$ is the matrix with the leading $r$ eigenvalues of $\Xcal \ttv{3} \bu^{(k)}$ on the diagonal. Standard eigensolvers return elements of $\bD$ alongside the eigenvectors so this approach does not change the complexity of our approach. For the theory given in Section \ref{sec:theory}, $d$ can be taken to be the $r$\textsuperscript{th} eigenvalue of $\bV_*\bV_*^T$ with minor modification.

\subsection{Deflation and Orthogonality} \label{sec:deflation}
While Algorithm \ref{alg:single_factor_sstpca} provides an efficient approach for estimating a single SS-TPCA factor, the multi-factor case is more difficult. The difficulty of the multi-factor case is a general characteristic of CP-type decompositions, posing significant computational and theoretical challenges, discussed at length by \citet{Han:2021} among others. To avoid these difficulties, we adapt a standard greedy (deflation) algorithm to the SS-TPCA setting; this sequential deflation approach to CP-type decompositions builds on well-known properties of the matrix power method and has been adapted to tensors independently by several authors, including \citet{Kolda:2005} and \citet{Allen:2012}.  While this approach was originally applied heuristically, \citet{Mu:2015} and \citet{Ge:2021} have shown conditions under which the deflation approach is able to recover an optimal solution. In our case, extending \citegenitive{Hotelling:1933} orginal approach to the tensor context yields the following successive deflation scheme:
%
\begin{algorithm}[H]
\begin{itemize}
\item Initialize: $\Xcal^1 = \Xcal$
\item For $k = 1, \dots, K$:
\begin{enumerate}
\item[(i)] Run Algorithm \ref{alg:single_factor_sstpca} on $\Xcal^k$ for rank $r_k$ to obtain ($\bu_k, \bV_k, d_k$)
\item[(ii)] Deflate $\Xcal^k$ to obtain $\Xcal^{k+1} = \Xcal^k - d_k \bV_k \circ \bV_k \circ \bu_k$
\end{enumerate}
\item Return $\left\{(\bu_k, \bV_k, d_k)\right\}_{k=1}^K$
\end{itemize}
\caption{Successive Deflation Algorithm for Multi-Factor $(r_1, \dots, r_K)$-SS-TPCA}
\label{alg:multi_factor_sstpca}
\end{algorithm}
As \citet{Mackey:2008} notes, classical (Hotelling) deflation fails to provide orthogonality guarantees when approximate eigenvectors are used, such as those arising from regularized variants of PCA. To address this, he proposes several additional deflation schemes with superior statistical properties. We discuss these alternate schemes in detail in Section \ref{refsec:deflation} of the Supplementary Materials. 

We note that the Algorithm \ref{alg:single_factor_sstpca} is only guaranteed to reduce the norm of the residuals and may not reduce the tensor rank at each iteration: in fact, in some circumstances, the tensor rank may actually be increased by our approach. We believe this is not a weakness of our approach: as signal is removed from a data tensor, the remaining residuals should increasingly resemble pure noise, which has high tensor rank almost surely \citep[Section 3.1]{Kolda:2009}. Similarly, as we remove estimated signal, the unexplained residual variance should decrease. While classical PCA does reduce the rank of the residual matrix at each iteration, this is an attractive but incidental property of linear algebra and not an essential statistical characteristic.

\subsection{Algorithmic Concerns}

The computational cost of Algorithm \ref{alg:single_factor_sstpca} is dominated by the eigendecomposition appearing in the $\bV$-update. Standard algorithms scale with complexity $\mathcal{O}(p^3)$, which can be reduced to $\mathcal{O}(p^2r)$ since we only need $r$ leading eigenvectors. Computation of the trace product $[\Xcal; \bV]$ is similarly expensive with complexity $\mathcal{O}(p^2Tr^2)$ but can be trivially parallelized across slices of $\Xcal$. Taken together, these give an overall per iteration complexity of $\mathcal{O}(p^2 \max\{Tr^2 / C, r\})$ when $C$ processing units are used. In practice, we have found the cost of the eigendecomposition to dominate in our experiments. As our theory below suggests, only a relatively small number of iterations, $I$, are required to achieve statistical convergence with $I$ growing logarithmically in the signal-to-noise ratio of the problem.

Algorithm \ref{alg:single_factor_sstpca} given above closely parallels the classical power method for computing singular value decompositions. As such, many techniques from that literature can be used to improve the performance of our approach or to adapt it to reflect additional computational constraints. While a full review of this literature is beyond the scope of this paper, we highlight the use of sketching techniques for particularly large data, one-pass techniques for streaming data, and distributed techniques \citep{Halko:2011,Tropp:2017,Tropp:2019,Li:2021}. When applied in the context of networks, it is particularly advantageous to take advantage of sparsity in $\Xcal$ in both the eigendecomposition and tensor algebra steps \citep{Phipps:2019}.

\subsection{Regularized SS-TPCA}

While the restriction to a rank-$k$ $\bV$ factor provides a powerful regularization effect as we will explore in the next section, it may sometimes be useful to regularize the $\bu$ term as well. For networks observed over time, a functional (smoothing) variant of SS-TPCA can be derived using the framework of \citet{Huang:2009}. Specifically, applying their smoothing perspective to the $\bu$-update of Algorithm \ref{alg:single_factor_sstpca}, the $\bu$-constraint becomes $\bu^T\bS_{\bu}\bu = 1$ for some smoothing matrix $\bS_{\bu} \succeq \bI$ yielding a modified $\bu$-update step: \[\bu^{(k+1)} = \frac{\bS_{\bu}^{-1}[\Xcal; \bV^{(k+1)}]}{\|\bS_{\bu}^{-1}[\Xcal; \bV^{(k+1)}]\|_{\bS_{\bu}}} = \frac{\bS_{\bu}^{-1}[\Xcal; \bV^{(k+1)}]}{\|[\Xcal; \bV^{(k+1)}]\|_{\bS_{\bu}^{-1}}}\]
where the $\bS_{\bu}$ norm is defined by $\|\bx\|_{\bS_{\bu}}^2 = \bx^T \bS_{\bu} \bx$. A similar approach was considered by \citet{Allen:2013} and \citet{Han:2021b} have recently established consistency of a related model. Alternatively, structured-sparsity in the $\bu$ factor can be achieved using a soft- or hard-thresholding step in the $\bu$-update, as discussed by \citet{Yuan:2013} and \citet{Ma:2013} respectively, or the $\ell_1$-based techniques considered by \citet{Witten:2009}. \citet{Allen:2019} discussed the possibility of simultaneously imposing smoothing and sparsity on estimated singular vectors and the core ideas of their approach could be applied to the $\bu$ vector as well, though we do not do so here.

\section{Consistency of SS-TPCA} \label{sec:theory}
Having introduced the SS-TPCA methodology, we now present a key consistency result for this decomposition. We analyze SS-TPCA under a tensor analogue of the spiked covariance model popularized by \citet{Johnstone:2001} or the low-rank mean model of \citet{Hoff:2007}. Specifically, we consider data generated with signal corresponding a rank-$k$ principal component $\bV_* \circ \bV_* = \bV_*\bV^T_*$ for some fixed $\bV_* \in \Vcal^{p \times k}$ and a loading vector $\bu_* \in \B_T$ and noise drawn from a  semi-symmetric sub-Gaussian tensor, $\Ecal$: 
\begin{equation}
    \Xcal = d\, \bV_* \circ \bV_* \circ \bu_* + \Ecal \label{eqn:tensor_spike_model}
\end{equation}
Here $d \in \R_{>0}$ is a measure of signal strength, roughly analogous to the square root of the leading eigenvalue of the spiked covariance model. In this scenario, we have the following consistency result: 
\begin{restatable}{theorem}{consistency} \label{thm:consistency}
Suppose $\Xcal$ is generated from the semi-symmetric model described above \eqref{eqn:tensor_spike_model} with elements of $\Ecal$ each independently $\sigma^2$-sub-Gaussian, subject to symmetry constraints. Suppose further that the initialization $\bu^{(0)}$ satisfies \[|1 - \langle \bu^{(0)}, \bu_*\rangle| \leq \tan^{-1}(0.5) - \frac{5 \ropnorm{\Ecal}}{d(1 - c)}\] for some arbitrary $c < 1$. Finally, assume $d \gtrsim r\sqrt{T} \sigma(\sqrt{p} + \sqrt{\log T})$. Then, the output of Algorithm \ref{alg:single_factor_sstpca} applied to $\Xcal$ satisfies the following
\[\min_{\epsilon = \pm 1} \|\epsilon \bu_* - \hat{\bu}\| / \sqrt{T} \lessapprox \frac{\sigma r \sqrt{p}}{d(1 - c)}.\]
with high probability. Here $\lesssim$ denotes an inequality holding up to a universal constant and $\lessapprox$ denotes an inequality holding up to a universal constant factor and a term scaling as $\sqrt{\log T}$.
\end{restatable}
A similar result holds for the $\bV$-factor: 
\begin{restatable}{theorem}{vconsistency} \label{thm:vconsistency}
Under the same conditions as Theorem \ref{thm:consistency}, the output of Algorithm \ref{alg:single_factor_sstpca} applied to $\Xcal$ satisfies, with high probability,
\begin{align*}
\min_{\bO \in \mathcal{V}^{r \times r}}\frac{\|\bV_* - \hat{\bV}\bO\|_2}{\sqrt{pr}} \lessapprox \frac{\sigma r \sqrt{T}}{d(1-c)}.
\end{align*}
\end{restatable}
We highlight that these results are first-order comparable to those for classical PCA attained by \citet{Nadler:2008}, though with more strenuous conditions on the initialization. (The results of the matrix power method almost surely do not depend on the choice of initialization, but this property is unique to the classical (unregularized) eigenproblem and does not hold in general.) A full proof of these results appears in the Supplementary Materials to this paper, but we provide a sketch of our approach in Section \ref{sec:proof_outline} below. A highlight of our approach is that it relies only on the Davis-Kahan theorem and standard concentration inequalities. While \citeauthor{Nadler:2008} uses sophisticated matrix perturbation results to establish higher-order consistency, we derive similar bounds for classical PCA using only the Davis-Kahan theorem in Section \ref{app:pca} of the Supplementary Materials. The fact that comparable results are obtained for the matrix and semi-symmetric tensor cases suggests that our results are essentially the best that can be obtained for this problem using elementary non-asymptotic techniques.

The initialization condition of Theorem \ref{thm:consistency} essentially assumes initialization within $53.1^{\circ} = \tan^{-1}(0.5)$ of the true $\bu_*$ factor, with additional accuracy needed to deal with higher-noise scenarios. In low-dimensions, this condition is easily satisfied, but it becomes more strenuous in the the large $T$ setting. We believe this condition to be, in part, an artifact of our proof technique: our experiments in Section \ref{sec:empirical} suggest that initialization in the correct orthant is typically sufficient. This is particularly compelling in the network setting, where it is reasonable to assume that all elements of $\bu$ are positive, \emph{i.e.}, that the network does not swap a large number of edges simultaneously, flipping the loading on the principal network. In situations where $\bu$ cannot be assumed positive, our approach is computationally efficient enough to allow for repeated random initializations.

We note that our results are stated for root mean squared error under sub-Gaussian noise for analytical simplicity, but that our approach could be used more broadly. In particular, if one assumes \textsc{iid} Bernoulli noise, corresponding to randomly corrupted edge indicators in the network setting, the results of \citet{Vu:2011} lead to significantly tighter bounds. In certain high-dimensional problems, elementwise error bounds on $\bu$ and $\bV$ may be more useful than an overall RMSE bound and could be derived using the $\ell_{\infty}$ eigenvector perturbation results of \citet{Fan:2018} or \citet{Damle:2020}. Finally, we note that because our analysis leverages Davis-Kahan (uniform) bounds, we do not assume any cancellation among elements of $\Ecal$, allowing our proofs to be applied in a dynamic adversarial setting, so long as $\ropnorm{\Ecal}$ remains bounded at each iteration. Replacing the uniform Davis-Kahan bounds with high-probability bounds such as those of \citet{ORourke:2018} could lead to tighter bounds or less stringent initialization and signal conditions.

\subsection{Connection with Regularized PCA} \label{sec:rank_pca}
At this point, the reader may wonder about the connection between traditional PCA and SS-TPCA: after all, our approach endows the space of semi-symmetric tensors with a real inner product, treating it as essentially Euclidean, and achieves the same estimation error as classical PCA under the appropriate rank-one model. As we now discuss, there is an equivalence between SS-TPCA and a certain form of regularized matrix PCA not previously considered in the literature. The regularization implicit in this equivalence is key to understanding how our method can perform well on large graphs and avoid the pitfalls of standard PCA in high-dimensional settings \citep{Johnstone:2009}.

In order to express a $p\times p \times T$-dimensional semi-symmetric tensor as a matrix, it is natural to form a $T \times \binom{p}{2}$ data matrix by vectorizing one triangle of each tensor slice and combining the vectorized slices as rows of a matrix. For a given semi-symmetric tensor, $\Xcal$, let us denote the corresponding data matrix as $\bX = \overline{\Mcal}_3(\Xcal)$ where $\overline{\Mcal}_k(\cdot)$ denotes tensor matricization of the upper triangle of each slice preserving the $k$\textsuperscript{th} mode, \emph{i.e.} $\bX_{i\cdot} = \uvec{\Xcal_{\cdot\cdot i}}$. Performing classical PCA on this matrix will have estimation error proportional to $\sigma \sqrt{\binom{p}{2}} / d \approx \sigma p / d$, which can be significantly worse than the $\sigma r \sqrt{p}/d$ rate attained by SS-TPCA for graphs having rank $r \ll \sqrt{p}$. Furthermore, classical PCA imposes no requirements on the estimated principal component and does not imply that the principal network given by $\unuvec{\hat{\bv}}$ has a low-rank structure.

\begin{table}[ht]
  \centering
    \begin{tabular}{c|c|cc}
      \toprule
      \multirow{2}{*}{Method} & \multirow{2}{*}{Data Dimension} & \multicolumn{2}{c}{Aligned RMSE}\\
      &&$\bu$-Factor & $\bv/\bV$-Factor \\
      \midrule
      Classical PCA & $T \times p$ & $\frac{\sigma \sqrt{p}}{d}$ & $\frac{\sigma \sqrt{T}}{d}$\\
      Matricization + Classical PCA & $T \times \binom{p}{2}$ &  $\frac{\sigma p}{d}$ & $\frac{\sigma \sqrt{T}}{d}$\\
      SS-TPCA (rank $1$) & $p \times p \times T$ & $\frac{\sigma \sqrt{p}}{d}$ & $\frac{\sigma \sqrt{T}}{d}$ \\
      SS-TPCA (rank $r$) & $p \times p \times T$ & $\frac{\sigma r \sqrt{p}}{d}$ & $\frac{\sigma r \sqrt{T}}{d}$ \\
      \bottomrule
    \end{tabular}
  \caption{Comparison of SS-TPCA with Matrix-Based Approaches. Under a ``signal plus noise'' model with signal strength $d$ and $\sigma^2$-sub-Gaussian noise, rank-1 SS-TPCA achieves the same accuracy as classical PCA on a $T \times p$ matrix. For the network PCA problem, the na\"ive approach of representing each observed network as a row vector and performing classical (matrix) PCA performs worse than SS-TPCA by a factor of $\sqrt{p}$. For large graphs well approximated by low-rank graphs, the improvements attained by SS-TPCA are particularly meaningful. Aligned RMSE refers to the scaled and rotated RMSE appearing in Theorems \ref{thm:consistency} and \ref{thm:vconsistency}, specifically $\text{ARMSE}(\bV, \hat{\bV}) = \min_{\bO \in \Vcal^{r \times r}} \|\bV - \hat{\bV}\bO\|_F / \sqrt{pr}$ for two matrices $\bV, \hat{\bV} \in \R^{p \times r}$. For vectors, \emph{e.g.}, the $\bu$-factor, aligned RMSE reduces to comparing $\bu$ against both $\hat{\bu}$ and $-\hat{\bu}$. Derivation of the error rates for classical PCA under a ``signal plus noise'' model appear in Section \ref{app:pca} of the Supplementary Materials.}
  \label{tab:my_label}
\end{table}

Low-rank structure of the principal network can be reimposed by adding an additional constraint on the right-singular vector: that is, by solving the regularized singular value problem \[\argmax_{\bu \in \overline{\B}^T, \bv \in \overline{\B}^{\binom{p}{2}}} \bu^T \overline{\Mcal}_3(\Xcal)\bv \quad \text{ subject to }  \rank\left(\unuvec{\bv}\right) = k.\] Finally, note that standard reformulation of the singular value problem as an eigenproblem via Hermitian dilation yields an equivalent regularized eigenvalue problem: 
\[\argmax_{\bx = (\bu, \bv) \in \overline{\B}^{T + \binom{p}{2}}} \bx^T\begin{pmatrix} \overline{\Mcal}_3(\Xcal)^T\overline{\Mcal}_3(\Xcal) & \bzero \\ \bzero & \overline{\Mcal}_3(\Xcal)\overline{\Mcal}_3(\Xcal)^T\end{pmatrix}\bx \quad \text{ subject to} \rank\left(\unuvec{\bx_{-(1:T)}}\right) = k\]
While these formulations are much more unwieldy than the tensor formulation, they reveal connections between our approach and the rich literature on sparse PCA. Specifically, our SS-TPCA Algorithm \ref{alg:single_factor_sstpca} can be interpreted as a variant of the truncated power method considered by \citet{Yuan:2013} and \citet{Ma:2013}, with the truncation to a $k$-sparse vector being replaced by truncation to a matrix with only $k$ non-zero eigenvalues. In this form, it is clear to see that \citegenitive{Yuan:2013} convergence analysis of the Truncated Power Method can be applied to give an alternate proof of the convergence of Algorithm \ref{alg:single_factor_sstpca}. 

While we do not consider minimax optimality in this paper, we note that substantially similar approaches have been shown to be minimax under comparable assumptions. In particular, \citet{Birnbaum:2013} show that classical PCA is rate-optimal under a dense signal model and that \citegenitive{Ma:2013} version of the truncated power method is optimal under a sparse signal model. Relatedly, \citet{Donoho:2014} establish optimality of a soft-thresholded SVD under the matrix version of the ``signal plus noise'' model. Because our method attains comparable convergence rates, we conjecture that SS-TPCA is minimax rate-optimal up to a logarithmic factor under conditions substantially similar to those of Theorem \ref{thm:consistency}.

This ``rank-unvec'' constraint appearing above has not previously appeared in the literature, but arises naturally from the low-rank structure random dot-product graphs. A similar constraint is implicit in certain unrelaxed formulations of the linear matrix sensing problem, but typically not analyzed as such \citep{Candes:2011b}.  Despite the non-convexity of this constraint, it is both computationally and theoretically tractable due to the nice properties of low-rank projections. We believe that this type of constraint can be applied more broadly in the analysis of network data, \emph{e.g.}, a variant of the network classification scheme considered by \citet{Relion:2019} with the coefficient matrix representing a low-rank graph rather than a sparse set of edges.

\subsection{Sketch of Proof of Theorems \ref{thm:consistency} and \ref{thm:vconsistency}} \label{sec:proof_outline}
In this section, we outline the proof of Theorems \ref{thm:consistency} and \ref{thm:vconsistency}. Full proofs of both results can be found in Section \ref{refsec:proofs} of the Supplementary Materials. Our analysis is of a tensor analogue of the classical power method for matrix decomposition; \citet{Hardt:2014} give an analysis of the matrix case that may provide useful background to our approach.

We first establish the following tail bound on the size of the noise tensor $\Ecal$ in terms of its semi-symmetric rank-$r$ operator norm: 
\begin{restatable}{proposition}{propropnorm} \label{prop:ropnorm} 
The semi-symmetric operator norm, $\ropnorm{\Ecal} = \max_{\bu, \bV \in \overline{\B}^T \times \Vcal^{p \times r}} |[\Ecal; \bV]^T\bu|$ can be deterministically bounded above by $r\sqrt{T}\max_i \lambda_{\max}(\Ecal_{\cdot\cdot i})$. Furthermore, if the elements of $\Ecal$ are independently $\sigma^2$-sub-Gaussian, subject to symmetry constraints, we have
\[\ropnorm{\Ecal} \leq cr\sqrt{T}\sigma\left(\sqrt{p} + \sqrt{\log T} + \delta\right)\] with probability at least $1 - 4e^{-\delta^2}$, for some absolute constant $c$.
\end{restatable}
The deterministic claim follows by bounding the elements of $[\Ecal; \bV]$ separately as $[\Ecal; \bV]_i = \sum_{j=1}^r \lambda_j(\Ecal_{\cdot\cdot i}) \leq r \lambda_{\max}(\Ecal_{\cdot\cdot i})$ and applying the standard result that $\|\bx\|_2 \leq \sqrt{T} \|\bx\|_{\infty}$ for any $\bx \in \R^T$. The probabilistic claim follows from standard bounds on the maximum eigenvalue of sub-Gaussian matrices and a union bound.

We combine the above result with a deterministic bound on the accuracy of Algorithm \ref{alg:single_factor_sstpca} for fixed $\Ecal$: 
\begin{restatable}{proposition}{deterministicconsistency_u} \label{prop:deterministicconsistency_u} 
Suppose $\Xcal = d\, \bV_* \circ \bV_* \circ \bu_* + \Ecal$ for a unit-norm $T$-vector $\bu_*$, a $p \times r$ orthogonal matrix $\bV_*$ satisfying $\bV_*^T\bV_* = \bI_{r \times r}$, $d \in \R_{\geq 0}$, and $\Ecal \in \R^{p \times p \times T}$ a semi-symmetric tensor. Then the result of Algorithm \ref{alg:single_factor_sstpca} applied to $\Xcal$ satisfies the following: 
\begin{align*}
  \min_{\epsilon = \pm 1} \|\epsilon \bu^* - \hat{\bu}\|_2 &\leq \frac{8\sqrt{2}\ropnorm{\Ecal}}{d(1 - c)}
\end{align*}
so long as $\ropnorm{\Ecal} < d$ and 
\[|1 - \langle \bu^{(0)}, \bu_*\rangle | \leq \tan^{-1}(0.5) - \frac{5 \ropnorm{\Ecal}}{d(1 - c)}\] for some arbitrary $c < 1$.
\end{restatable}
Combining Propositions \ref{prop:ropnorm} and \ref{prop:deterministicconsistency_u} yields Theorem \ref{thm:consistency}. The proof of Theorem \ref{thm:vconsistency} is virtually identical.

We establish Proposition \ref{prop:deterministicconsistency_u} using an iterative analysis of Algorithm \ref{alg:single_factor_sstpca} which controls the error in $\bV^{(k)}$ in terms of $\bV^{(k-1)}$ and repeats this to convergence. Applying \citegenitive{Yu:2015} variant of the Davis-Kahan theorem to the $\bV$-update, we find 
\begin{align*}
    \|\sin \Theta(\bV^*, \bV^{(k+1)})\|_F \leq 2 \sqrt{r}|1 - \cos \angle(\bu_*, \bu^{(k)})| + 2\sqrt{r} \ropnorm{\Ecal}{d}.
\end{align*} Similarly, applying the Davis-Kahan theorem to the $\bu$-update yields 
\begin{align*}
    |\sin \angle(\bu_*, \bu^{(k+1)})| &\leq 2|1 - \langle \bV_*, \bV^{(k+1)}\rangle^4| + 4r\frac{\ropnorm{\Ecal}}{d} + 2r^2\frac{\ropnorm{\Ecal}^2}{d^2}
\end{align*} Due to the non-linear $\textsf{Norm}(\cdot)$ function, this is not as simple as the $\bV$-update and requires applying the Davis-Kahan theorem to the matrix pair $\tilde{\bu}^{(k+1)} \circ \tilde{\bu}^{(k+1)}$ and $d^2 \bu_* \circ \bu_*$, where $\tilde{\bu}^{(k+1)} = [\Xcal; \bV^{(k+1)}]$ is the pre-normalized iterate. Because $\bu^{(k+1)}$ is an eigenvector of $\tilde{\bu}^{(k+1)} \circ \tilde{\bu}^{(k+1)}$ by construction, the eigenvector bound provided by Davis-Kahan is exactly what is needed to control $\theta_{\bu_{k+1}} = \angle(\bu_*, \bu^{(k+1)})$. 

With these two bounds in hand, we can bound the sine of the $\bV$-angle in terms of the cosine of the $\bu$-angle and \emph{vice versa}. To connect these, it suffices to assume we are in the range of angles such that $|\sin \theta| \geq 2|1 - \cos \theta|$ or $\theta \in [0, \tan^{-1}(0.5)) \approx [0, 53.1^{\circ})$. In order to ensure a substantial contraction at each step, we make the slightly stronger assumption that
\begin{align*}
   2|1 - \cos \theta_{\bu_k}| \leq c_u |\sin \theta_{\bu_k}| \quad \text{ and } \quad 
   2|1 - \cos \theta_{\bv_k}| \leq c_v |\sin \theta_{\bv_k}|
\end{align*}
for all $k$ and for some fixed $c_u, c_v < 1$.  From here, we iterate our $\bu$- and $\bV$-update bounds to show that the error in the $\bu$-iterates is bounded above by 
\[|\sin \theta_{\bu_k}| \leq c^k |\sin \theta_{\bu^{(0)}}| + \frac{8 \ropnorm{\Ecal}}{d} \sum_{i=0}^{k-1}c^i\]
where $c = c_uc_v$. Taking the $k \to \infty$ limit, this gives the error bound $\frac{8\ropnorm{\Ecal}}{d(1 - c)}$ for the final estimate $\hat{\bu} = \lim_{k\to\infty} \bu^{(k)}$. Finally, we note that given the geometric series structure of the above bound, only a logarithmically small number of iterations are required for the iterates to converge up to the noise bound of the data.

An important technical element of our proof is to ensure that the iterates remain within the strict $(c_u, c_v)$-contraction region for all $k$.  While this follows immediately for the noiseless case, the noisy case is more subtle, requiring us to balance the non-expansive error from our initialization with the effect of the noise $\Ecal$ which recurs at each iteration. To ensure this, we need our iterates to be bounded away from the boundary of the contraction region so that the sequence of ``contract + add noise'' does not increase the total error at any iterate. We term this non-expansive region the ``stable interior'' of the contraction region. Simple algebra shows that assuming assuming $\theta_{\bu_k}$ is in the stable interior, \emph{i.e.}, \[|1 - \cos \theta_{\bu_k}| \leq \tan^{-1}(0.5) - \frac{5 \ropnorm{\Ecal}}{d(1 - c_uc_v)},\] implies that $\theta_{\bu_{k+1}}$ is in the stable interior so it suffices to make this stronger assumption on the initialization $\theta_{\bu^{(0)}}$ only. This style of algorithm-structured analysis has recently found use in the analysis of non-convex problems: notable examples include recent work by \citet{Fan:2018b}, \citet{Sun:2016} and \citet{Zhao:2018}. In particular, the contraction assumptions above parallel the \emph{Restricted Correlated Gradient} condition of \citegenitive{Han:2022} low Tucker-rank framework. Compared with those papers, our result holds under much more general assumptions, only requiring bounds on the signal-to-noise ratio of the problem and a loose bound on initialization. Specifically, we do not require standard high-dimensional assumptions of sparsity, restricted strong convexity, or incoherence.

\section{Empirical Results} \label{sec:empirical}

\begin{figure}[htb]
\centering
\includegraphics[width=\textwidth]{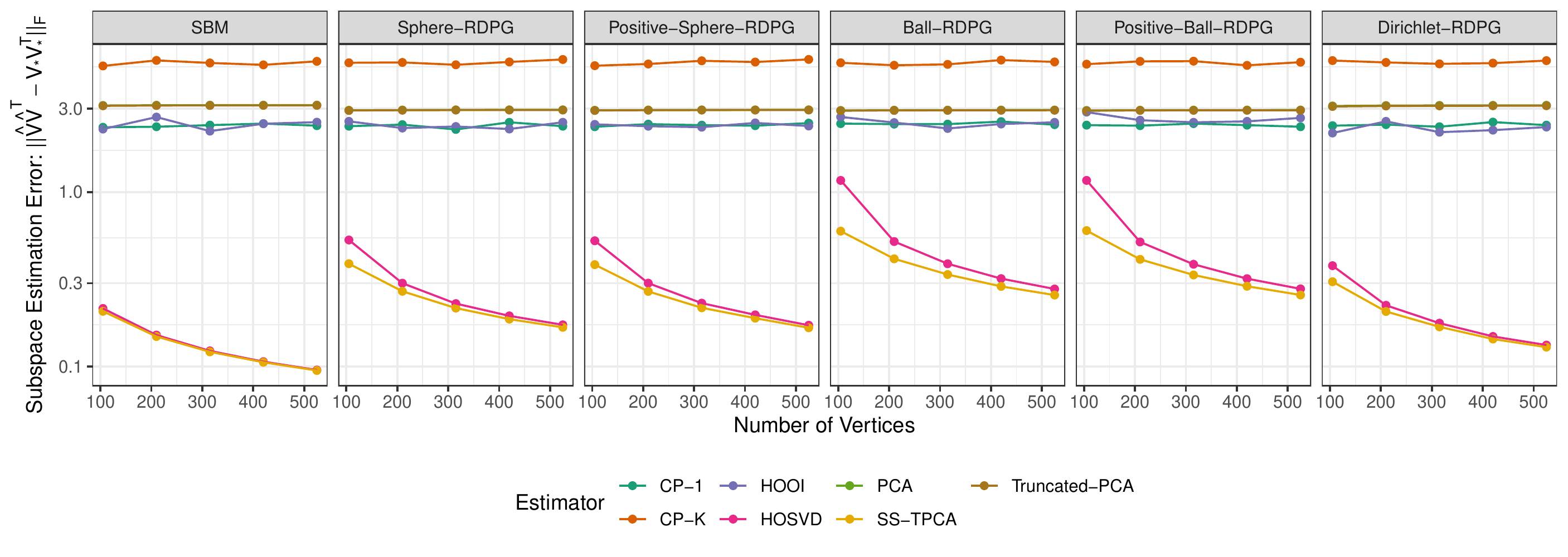}
\caption{Comparison of low-rank subspace estimation under different network PCA models, under $T = 20$ \textsc{iid} samples from rank-5 graphs of varying size. SS-TPCA (gold) outperforms competitor methods, with the HOSVD-based Tucker method (pink) also performing well. Matricization-based approaches, both with and without an additional truncation step, and CP-ALS approaches are not competitive. SBM samples are generated with intra-group connection probability $p = 0.8$ and inter-group connection probability $q = 0.2$. Dirichlet-RDPG refers to latent positions sampled from the 5-dimensional simplex using a Dirichlet distribution with parameter $\balpha = 0.3 * \bone_5$. Samples are generated and tensor decompositions are performed using the \textsf{igraph} and \textsf{rTensor} packages respectively. The surprisingly competitive performance of the HOSVD seems to be related to the regularized PCA formulation of our model as discussed in Section \ref{sec:rank_pca}.}
\label{fig:comparison}
\end{figure}
\subsection{Simulation Studies}
In this section, we demonstrate the effectiveness of SS-TPCA under various conditions and give empirical evidence to support the theoretical claims of the previous two sections. Specifically, we seek to demonstrate the following four claims empirically: i) SS-TPCA outperforms classical PCA methods and generic tensor decompositions in the analysis of network valued data; ii) SS-TPCA exhibits rapid \emph{computational} convergence to a region surrounding the true parameter value; iii) SS-TPCA exhibits classical \emph{statistical} convergence rates as $T$ and $p$ vary; and iv) SS-TPCA is robust to the choice of initialization, $\bu^{(0)}$.

\paragraph{Simulation I - Methods Comparison:} We generate $\Xcal$ using \textsc{iid} samples from different variants of the Random Dot Product Graph framework, including a stochastic block model and latent position graphs on the simplex, unit sphere, and unit ball. In each case, we simulate $T = 20$ graphs with a rank-5 structure, but vary the number of nodes from 105 to 525. We compare SS-TPCA with several potential competitors, including classical (vectorized) PCA, classical PCA followed by a low-rank truncation step, a rank-1 CP decomposition, a rank-5 CP decomposition, and a Tucker decomposition of order (5, 5, 1). The CP decompositions are estimated using alternating least squares, while we consider both the Higher-Order SVD (HOSVD) and the Higher-Order Orthogonal Iterations (HOOI) approaches to estimating Tucker decompositions. See \citet{Kolda:2009} for additional background. Our results appear in Figure \ref{fig:comparison}, where we report the subspace estimation error for each method. (We do not report estimation error for the $\bu$-factor as it is uninteresting for \textsc{iid} samples, but all methods estimate both terms.) SS-TPCA consistently outperforms other estimators, with the HOSVD-estimated Tucker decomposition also performing well.

\begin{figure}[htb]
\centering
\includegraphics[width=0.5\textwidth]{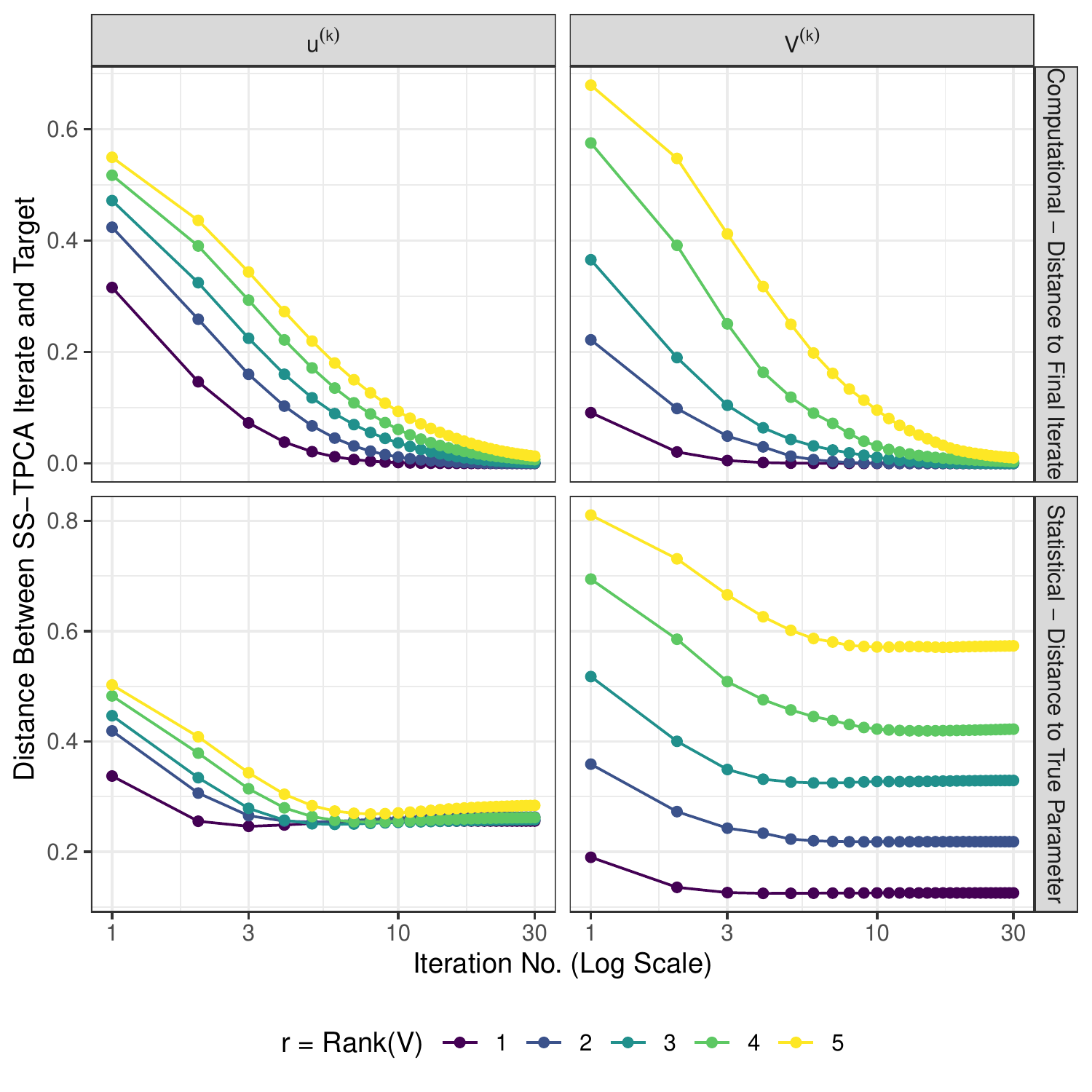}
\caption{Comparison of the \emph{computational} and \emph{statistical} convergence of Algorithm \ref{alg:single_factor_sstpca}. While SS-TPCA takes approximately 30 iterations to converge to $(\hat{\bu}, \hat{\bV})$, the intermediate iterates $(\bu^{(k)}, \bV^{(k)})$ have essentially the same statistical accuracy after $k = 4$ iterates.}
\label{fig:comp_convergence}
\end{figure}

\paragraph{Simulation II - Computational and Statistical Convergence:} We generate $\Xcal$ from the model from the model $\Xcal = d\, \bV_* \circ \bV_* \circ \bu_* + \Ecal$, with $T = 20$ observations and $p = 200$ vertices.  Because our main theorems do not place any restrictions on $\bV_*$, we draw $\bV_*$ uniformly at random from the rank-$r$ Stiefel manifold. We fix $\bu_* = \bone_T / \sqrt{T}$ and initialize Algorithm \ref{alg:single_factor_sstpca} with $\bu^{(0)}$ drawn randomly from the set of positive unit vectors. Finally, we generate $\Ecal$ such that each slice of $\Ecal$ is an independent draw from the Gaussian orthogonal ensemble (GOE): that is, the off-diagonal elements are drawn from independent standard normal distributions and the diagonal elements are independent samples from a $\mathcal{N}(0, 2)$ distribution. We vary the signal strength $d = 15 r^{-1/4}$ to preserve a signal to noise ratio of slightly more than one (note $\sqrt{p} \approx 14.14$). Figure \ref{fig:comp_convergence} depicts the convergence of the iterates of Algorithm \ref{alg:single_factor_sstpca} under this regime. It is clear that the iterates of Algorithm \ref{alg:single_factor_sstpca} achieve \emph{statistical convergence}, in the sense of non-decreasing estimation error, in approximately one tenth the iterations needed to achieve computational convergence.

\begin{figure}[htb]
\centering
\includegraphics[width=\textwidth]{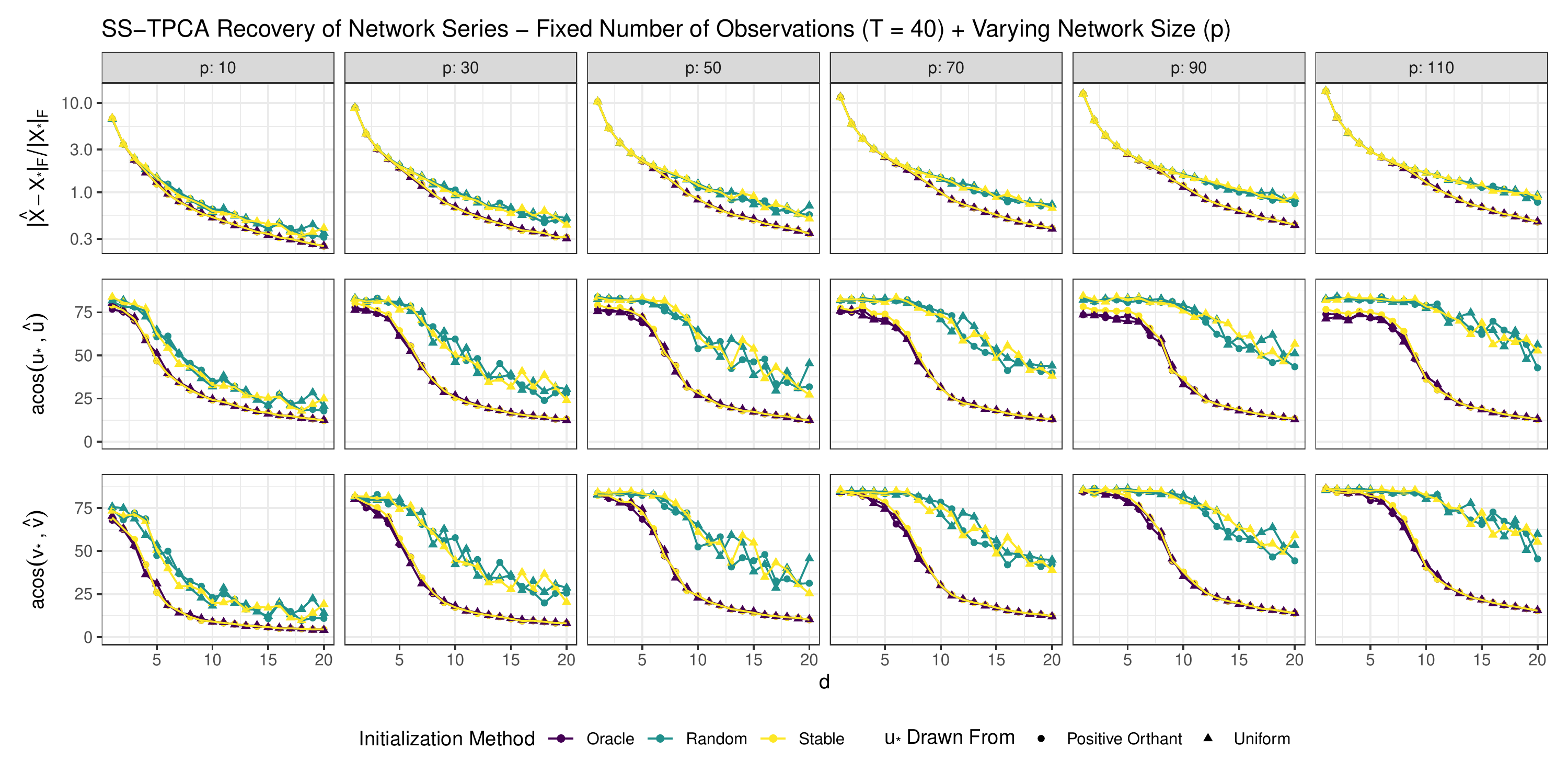}
\caption{Performance of Algorithm \ref{alg:single_factor_sstpca} for recovery of a rank-1 semi-symmetric tensor of norm $d$ with added noise from a Gaussian Orthogonal Ensemble. Tensor recovery is accurate across several measures including the normalized reconstruction error between $\hat{\Xcal}$ from Algorithm \ref{alg:single_factor_sstpca} and the signal tensor $\Xcal_* = d\, \bV_* \circ \bV_* \circ \bu_*$ (top row), the angle between $\bu_*$ and $\hat{\bu}$ (middle row), and the angle between $\bV_*$ and $\hat{\bV}$ (bottom row). As expected, oracle initialization (purple, $\bu_0 = \bu_*$) consistently performs the best of the initialization schemes considered, but random initailization (green) and ``stable'' initialization (yellow, to the all-ones vector) also perform well. In scenarios where $\bu_*$ is restricted to have all positive elements, stable initialization performs almost indistinguishably from oracle initialization; in scenarios where $\bu_*$ is allowed to have both positive and negative elements, stable initialization performs essentially as well as random initialization, suggesting that our method is quite robust to the specific choice of initialization. We fix $T = 40$ observations and vary the dimension of $\bV_*$ from 10 (far left column) to 110 (far right column); as would be expected, performance does decay for larger dimensions, but the decay is rather slow, consistent with our theoretical results.}
\label{fig:recovery_fixed_T}
\end{figure}

\begin{figure}[htb]
\centering
\includegraphics[width=\textwidth]{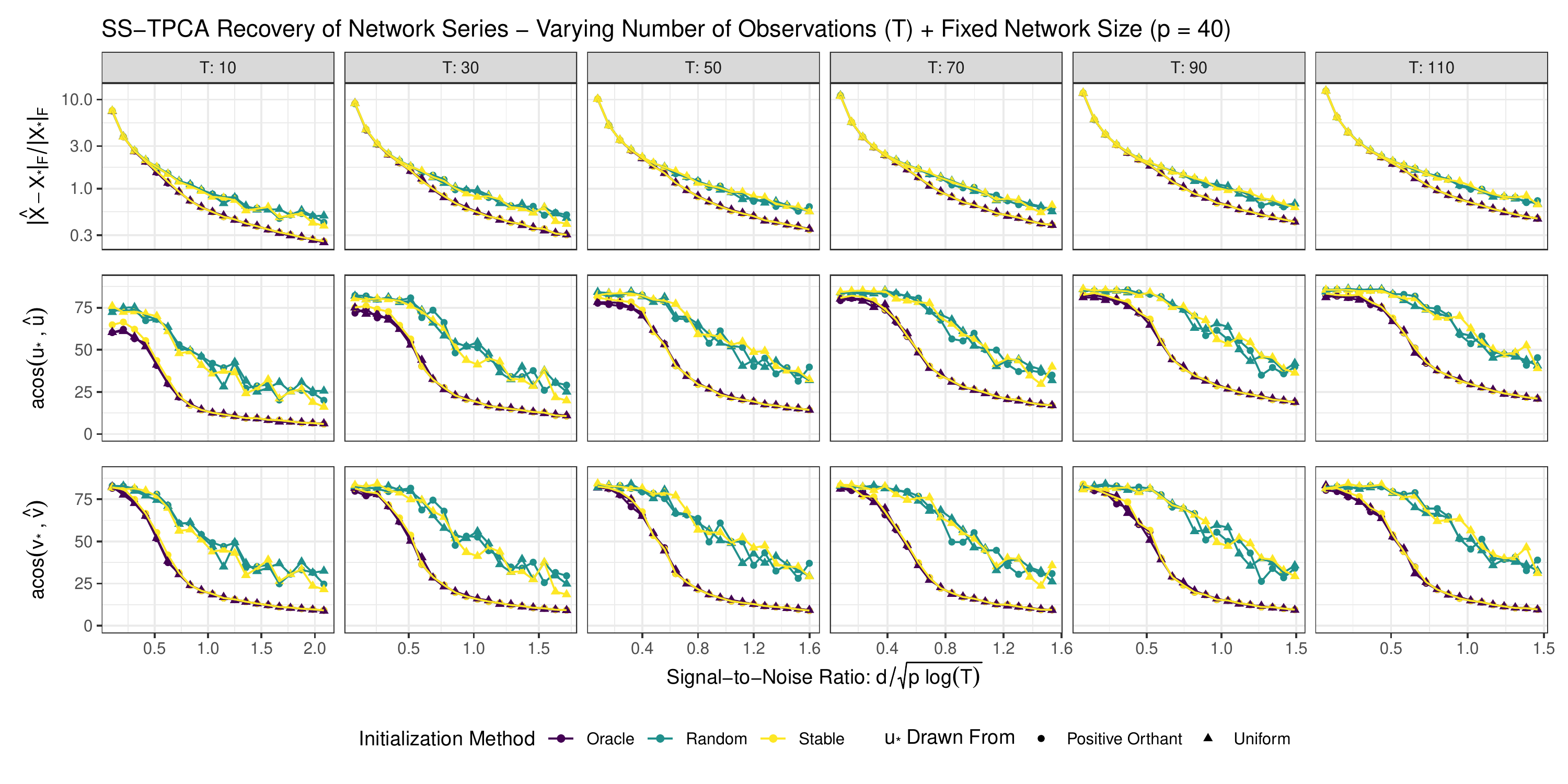}
\caption{Performance of Algorithm \ref{alg:single_factor_sstpca} for recovery of a rank-1 semi-symmetric tensor of norm $d$ with added noise from a Gaussian Orthogonal Ensemble. Under the same generative model as Figure \ref{fig:recovery_fixed_T}, we here fix the dimension of $\bV_*$ at 40 and vary the number of observations $T$ from $T = 10$ (far left) to $T = 110$ (far right). Here we rescale the $x$-axis to the effective signal-to-noise ratio $d / (\sqrt{p \log T})$ and note that accuracy to within $25^{\circ}$ can be obtained even in relatively difficult $\text{SNR}\approx 1$ scenarios. (For $p = 40$, $\angle(\hat{\bV}, \bV_*) \approx 25^{\circ}$ corresponds to an $\ell_2$ error of  $\sqrt{2}\sin(25^{\circ}) \approx 0.6$ or a root mean squared error $0.1$ across all elements of $\hat{\bV}$. This is well within the bounds of usability for many network science applications where the elements of $\Xcal$ take $\{0, 1\}$-values.}
\label{fig:recovery_fixed_p}
\end{figure}

\paragraph{Simulation III - Convergence Rates and Initialization Robustness:} We generate $\Xcal$ as before, but fix $r = 1$ so $\bV_*$ is drawn uniformly at random from the rank-1 Stiefel manifold, \emph{i.e.}, the unit sphere. We now also let $\bu_*$ be generated randomly under two different mechanisms: $\bu_*$ drawn uniformly at random from the unit sphere and $\bu_*$ drawn uniformly at random from the portion of the unit sphere in the first (positive) orthant, \emph{i.e.}, non-negative unit vectors. As before, each slice of $\Ecal$ is independently drawn from the GOE. From well-known bounds on the GOE, we have $\opnorm{\Ecal} \approx \sqrt{p \log T}$, giving an effective signal-to-noise ratio of $d / \sqrt{p \log T}$ for all experiments. We consider three approaches to initialize Algorithm \ref{alg:single_factor_sstpca}: i) oracle initialization - $\bu^{(0)} = \bu_*$; ii) random initialization where $\bu^{(0)}$ is drawn at random from the unit sphere; and iii) ``stable'' initialization with $\bu^{(0)} = \bone_T / \sqrt{T}$. 

Figure \ref{fig:recovery_fixed_T} shows the results of our study in the case of fixed $T$ (number of samples) and varying $p$ (network size) from 10 to 110. We present three measures of accuracy: i) the normalized reconstruction error between $\hat{\Xcal} = \hat{d}\, \hat{\bV} \circ \hat{\bV} \circ \hat{\bu}$ and the signal tensor $\Xcal_* = d\, \bV_* \circ \bV_* \circ \bu_*$; ii) the angle between $\bu_*$ and $\hat{\bu}$; and iii) the angle between the subspaces spanned by $\bV_*$and $\hat{\bV}$, calculated as $\cos^{-1}\left[\sigma_{\min}(\bV_*^T\hat{\bV})\right]$, where $\sigma_{\min}(\cdot)$ denotes the last (non-zero) singular value. Note that, for very accurate recovery, the $\cos^{-1}(\cdot)$ used in computing the angle introduces a non-linear artifact: Figures \ref{fig:comparison} and \ref{fig:comp_convergence}, as well as $\hat{\Xcal}$, depict the expected parametric ($n^{-1/2}$) convergence rate. Consistent with Theorems \ref{thm:consistency} and \ref{thm:vconsistency}, we observe errors that decay rapidly in $d$ for all three measures under all generative and initialization schemes. Comparing the case with $\bu_*$ from the positive orthant or the entire unit sphere, we observe that random initialization does well in both cases, with accuracy only slightly worse than oracle initialization, and that stable initialization achieves essentially the same performance as oracle initialization for positive $\bu_*$. 

Figure \ref{fig:recovery_fixed_p} presents similar results taking the network size of $p = 40$ nodes fixed and varying $T$, though here with the effective signal-to-noise ratio $d / \sqrt{p \log T}$ as the ordinate ($x$-axis) rather than the scale factor $d$. As in Figure \ref{fig:recovery_fixed_T}, the results of Theorems \ref{thm:consistency} and \ref{thm:vconsistency} are confirmed, with error decaying quickly and accurate recovery even at signal-to-noise ratios of approximately 1, and with both random and stable initializations being qualitative competitive with oracle initialization and with the stable initialization being essentially indistinguishable from the oracle scheme for positive $\bu_*$. Taken together, these two experiments demonstrate that Algorithm \ref{alg:single_factor_sstpca} is robust to the initialization scheme and consistently performs well, even in situations where the initialization and signal strength requirements of our theorems are violated. 



\subsection{Case Study: Voting Patterns of the Supreme Court of the United States} \label{sec:application_scotus}
\begin{figure*}[t]
\centering
\includegraphics[width=\textwidth]{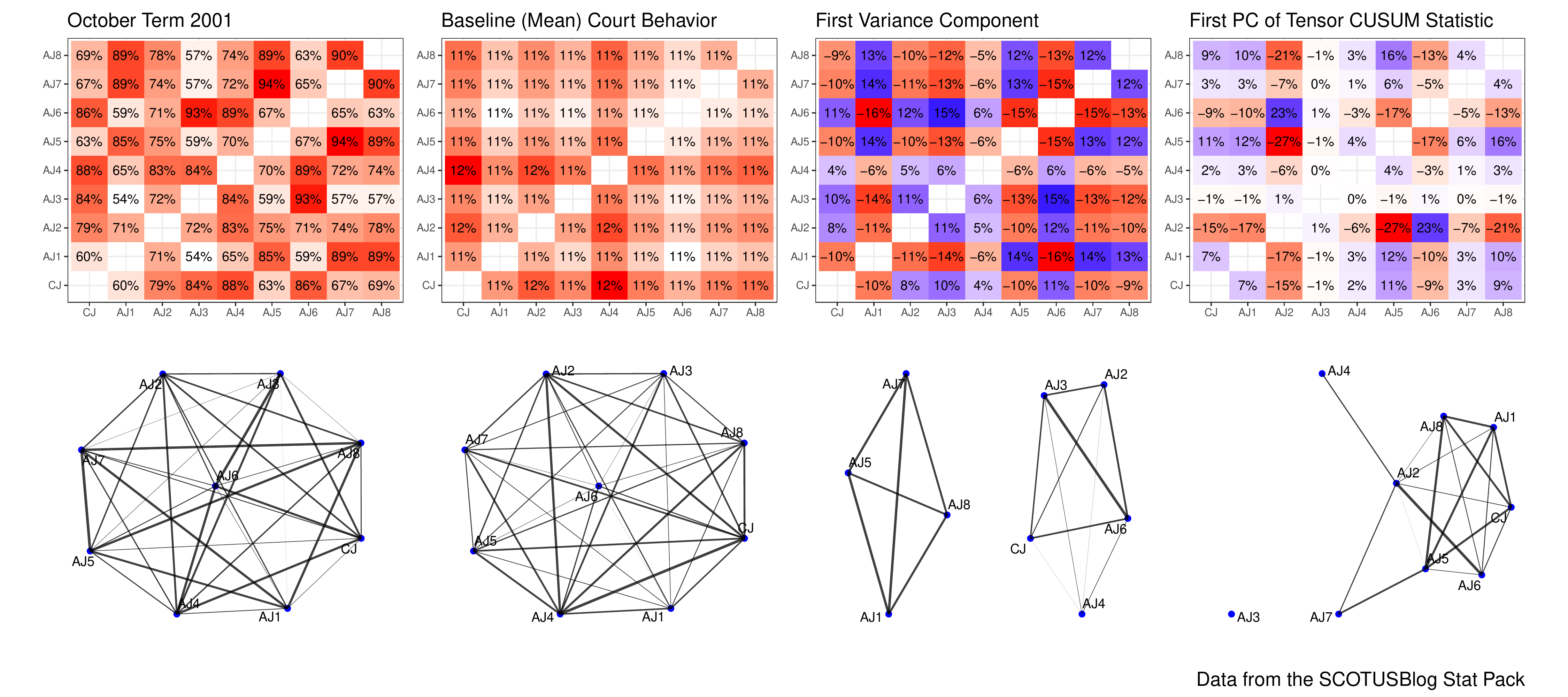}
\caption{Application of SS-TPCA to the Supreme Court voting example described in Section \ref{sec:application_scotus}. OT 2001 (left column) depicts a typical and relatively non-controversial \scotus term, with the raw agreement data (top row) and implied agreement network (bottom row) indicating a high degree of concurrence among the justices. The principal network identified by Network PCA (second column) is a nearly ``flat'' fully connected network, signifying the high fraction of low-profile but nearly unanimous cases that dominate \scotus's docket. After extracting this factor, the major mode of variation (third column) away from baseline unanimity is a conservative (seats CJ, AJ2, AJ3, AJ4, and AJ6)-liberal (seats AJ1, AJ5, AJ7, AJ8) split among the justices, typical of high-profile ``5-4'' decisions. Finally, the combination of \cusum analysis with SS-TPCA (right column) identifies the retirement of the moderate Justice O'Connor (seat AJ2) as the most important shift in the OT 1995-OT 2020 time period; the $\hat{\bu}$ factor from this analysis identifies OT 2005 as the most likely site of the change. Justice O'Connor retired in the middle of OT 2005, suggesting that her successor, Justice Alito, quickly established himself as a significantly more conservative justice than his predecessor. Judicial agreement data for this analysis was extracted from the annual ``stat packs'' prepared by the legal news and commentary site SCOTUSBlog (\url{scotusblog.com}).}
\label{fig:scotus}
\end{figure*}

Finally, we apply SS-TPCA to analyze voting patterns of the justices of the Supreme Court of the United States (\scotus) from a period from October 1995 to June 2021. Specifically, we construct a series of $9 \times 9$ networks with nodes corresponding to each ``seat'' on the court: that is, a justice and her successor are assigned to the same node even if they have substantially different judicial philosophies, \emph{e.g.}, Ruth Bader Ginsberg and Amy Coney Barrett. Edge weights for each network are given by the fraction of cases in which the pair of justices agree in the specific judgement of the court: we ignore additional subtleties that arise when two justices agree on the outcome, but disagree on the legal reasoning used to reach that conclusion. We repeat this process for 25 year-long terms beginning in October 1995 (``October Term 1995'' or ``OT 1995'') to the term ending in June 2021 (``OT 2020''). Our results are presented in Figure \ref{fig:scotus}.

We begin by finding the principal (mean) network from this data by applying Algorithm \ref{alg:single_factor_sstpca} to $\Xcal$ directly. This yields a rank-1 principal component with $\hat{\bV} \approx \bone * 1/3$ or, equivalently, $\hat{\bV} \circ \hat{\bV} \approx \bJ_{9 \times 9} * 1/9$, where $\bJ$ is the all-ones matrix. This suggests that the baseline behavior of \scotus is a broad-based consensus, typified by unanimous or nearly unanimous decisions. While this is somewhat at odds with the popular perception of \scotus and news coverage highlighting controversial decisions, it is consistent with the fact that the majority of \scotus cases are focused on rather narrow questions of legal and procedural arcana on which the justices can find broad agreement. We also note that the justices have discretion over the majority of their docket and may choose to favor cases where unanimity is likely. 

We next apply SS-TPCA to the residuals from the mean analysis in order to identify the typical patterns of agreement after removing the unanimity factor. In this case, we find a principal network with two clear components of 5 and 4 vertices: a closer examination of this split reveals a stereotypical conservative (5) - liberal (4) split, where justices tend to vote in agreement with other justices nominated by a president from the same political party. This factor thus clearly identifies the partisan divide highlighted in media coverage and popular perceptions of \scotus. We note, that the signal of the first factor ($\hat{d}_1 \approx 35$) is much stronger than the signal of the second factor ($\hat{d}_2 \approx 4.8$) suggesting that public perception is driven by a small number of high-profile and divisive cases.

Finally, we note that SS-TPCA techniques can be used to identify change points in a time-ordered network series under a rank-$r$ mean-shift model. Specifically, we adapt a proposal of \citet{Wang:2018-Changepoint} to the network setting and perform SS-TPCA on the Cumulative Sum Control Chart (``\cusum'') tensor $\Ccal$ given by
\[\Ccal_{ijt} = \sqrt{\frac{T}{t(T - t)}}\left(\frac{t}{T}\sum_{\tau = 1}^t \Xcal_{ij\tau} - \sum_{\tau = 1}^T \Xcal_{ij\tau}\right) \]
By construction, $\Ccal \in \R^{p \times p \times (T - 1)}$ inherits the semi-symmetric structure of $\Xcal$ and so can be used as input to Algorithm \ref{alg:single_factor_sstpca} with rank $r$. The resulting ``time factor,'' $\hat{\bu}$, can then be used for change point analysis with $\hat{\tau} = \argmax_{1 \leq \tau \leq (T - 1)} |\hat{u}_{\tau}|$ being the most likely change point. A combination of Theorem \ref{thm:consistency} and standard change point detection theory implies that this approach correctly identifies a change point from data generated with mean $\bV_*^{(1)}$ to $\bV_*^{(2)}$ with probability inversely proportional to the effective signal-to-noise ratio \[\opnorm{\bV_*^{(1)} - \bV_*^{(2)}}\sqrt{\min\{\tau^{(1)}, \tau^{(2)}\}}/\sigma\]
where $\tau^{(1)}$ and $\tau^{(2)}$ are the number of observations before and after the change point directly. This compares well with optimal rates for the univariate \cusum model \citep{Wang:2020} and with recent work on network change point detection \citep{Wang:2021}, specializing the latter results to the low-dimensional (non-sparse) regime. 

We apply this \cusum type analysis to identify major changes in the voting patterns of \scotus justices. The results of this analysis suggest that the most important change, estimated by the largest element of $\hat{\bu}$, occurred in OT 2005 and was driven by the replacement of Justice Sandra Day O'Connor, a moderate conservative who would occasionally vote with her more liberal colleague Justice Souter, by Justice Samuel Alito, a firm conservative who almost always votes in agreement with fellow conservative Justice Thomas. The importance of this shift is commonly noted by legal commentators. We note that, while the replacement of Justice Ruth Bader Ginsberg by Justice Amy Coney Barett is likely to be even more important in the overall history of the court, Justice Barrett only served for one \scotus term in our sample and hence cannot be identified by change point analyses. Further, we also note that \cusum analysis suggests essentially no change in voting behavior when Justice Neil Gorsuch replaced Justice Antonin Scalia, consistent with their similar judicial philisophies and Justice Gorsuch's adherence to judicial philisophy originally expounded by Justice Scalia.

Note that the networks in this case study are quite small, with only nine vertices, and barely cross into the high-dimensional regime with $\binom{9}{2} = 36$ distinct edges for $25$ observations. In this case, classical PCA on edge weights obtains results that are qualitatively similar to those presented here, as the effect of low-rank regularization is minimal. An additional case study in Section \ref{app:empirical} of the Supplementary Materials analyzes correlation networks of international stock markets.
\section{Discussion} \label{sec:discussion}
We have presented a novel framework for multivariate analysis of network-valued data based on a semi-symmetric tensor decomposition we term SS-TPCA. Our major contribution is a theoretical analysis of SS-TPCA, where we show that SS-TPCA achieves estimation accuracy comparable to classical PCA, despite the significantly higher dimensionality of the tensor problem. We also highlight connections between SS-TPCA and a novel form of regularized PCA, based on a ``rank-unvec'' constraint. This constraint naturally captures a wide range of random graph models and can be applied in other supervised and unsupervised learning problems. Our analysis has three attractive properties: i) it relies only on elementary tools of matrix perturbation theory; ii) it establishes consistency under a very general noise model, allowing for the possibility of adversarial noise;  and iii) by analyzing a specific computational approach, it avoids considerations of global versus local optimality. Given this combination of ease and power, we expect that our analytic approach will find use elsewhere. Finally, we demonstrate through simulation that our method is robust to choice of initialization and outperforms other methods in low- and high-signal regimes. 

We expect several extensions of our work to be of particular interest: given the large scale of much network data, streaming, parallel, and approximate versions of Algorithm \ref{alg:single_factor_sstpca} will broaden the applicability of SS-TPCA. Network time series also are typically highly autocorrelated and regularization of the $\bu$-term reflecting this fact may significantly improve performance. While the low-rank model at the heart of SS-TPCA captures a wide range of useful graph models, analogues of our work for directed, preferential attachment, multi-edge, and other graph structures would also be quite helpful in certain application domains. Robust versions of SS-TPCA, building on similar techniques for matrix PCA, are also of interest. Finally, we have assumed that the set of vertices is fixed and consistent node labels are available: unfortunately, this is rarely the case for large-scale networks arising from social media, telecommunications, or other important domains. Node sets change from one day to the next as users create and delete accounts and, despite recent progress, the graph alignment problems resulting from unlabeled nodes remain computationally prohibitive. Recent development in the theory of graphons suggests the use of continuous (functional) representations of graphs of different sizes and we are excited to pursue this avenue in future work. 

\section*{Acknowledgements}
MW's research is supported by an appointment to the Intelligence Community Postdoctoral Research Fellowship Program at the University of Florida Informatics Institute, administered by Oak Ridge Institute for Science and Education through an interagency agreement between the U.S. Department of Energy and the Office of the Director of National Intelligence. GM's research is supported by NSF/DMS grant 1821220.

\printbibliography
\end{refsection}

\clearpage
\appendix 
\setcounter{figure}{0}
\renewcommand{\thefigure}{A\arabic{figure}}
\setcounter{table}{0}
\renewcommand{\thetable}{A\arabic{table}}

\renewcommand*{\theappendixtheorem}{\thesection.\arabic{appendixtheorem}}
\renewcommand*{\theappendixproposition}{\thesection.\arabic{appendixproposition}}
\renewcommand*{\theappendixremark}{\thesection.\arabic{appendixremark}}

\begin{refsection}
\begin{center}{\LARGE \bf Supplementary Materials}\end{center}
\section{Additional Discussion for Subsection \ref{sec:deflation} - \nameref{sec:deflation}} \label{refsec:deflation}

As \citet{Mackey:2008} notes, classical (Hotelling) deflation fails to provide orthogonality guarantees when approximate eigenvectors are used, such as those arising from regularized variants of PCA. To address this shortcoming, he proposes ``projection deflation'' and ``Schur deflation'' techniques which provide stronger orthogonality guarantees and consequently improve the estimation of subsequent components. Because we cannot treat the SS-TPCA components estimated by Algorithm \ref{alg:single_factor_sstpca} as true eigenvectors, we propose three distinct tensor deflation schemes, each providing distinct degrees of orthogonality:
\begin{align}
    \Xcal^{k+1} &= \Xcal^{k} - d_k \bV_k \circ \bV_k \circ \bu_k \tag{HD} \label{eq:hd}\\
    \Xcal^{k+1} &= \Xcal^{k} \ttm{1} [\bI_{p \times p} - \bV_k \circ \bV_k] \ttm{2} [\bI_{p \times p} - \bV_k \circ \bV_k] \ttm{3} [\bI_{T \times T} - \bu_k\bu_k^T] \label{eq:pd}\tag{PD} \\
    \begin{split}\Xcal^{k+1} &= \widetilde{\Xcal}^{k+1} \ttm{3} [\bI_{T \times T} - \bu_k\bu_k^T]  \\
    \text{ where } \widetilde{\Xcal}^{k+1}_{\cdot\cdot i} &= \Xcal^k_{\cdot\cdot i} - \Xcal^k_{\cdot\cdot i}\bV_k(\bV_k^T\Xcal^k_{\cdot\cdot i}\bV_k)^{-1}\bV_k^T \Xcal^k_{\cdot\cdot i} \text{ for all } 1 \leq i \leq T \end{split}\tag{SD} \label{eq:sd}
\end{align}
The Hotelling deflation \eqref{eq:hd} and projection deflation \eqref{eq:pd} schemes are tensor extensions of the analogous results for non-symmetric matrix decompositions given by \citet{Weylandt:2019c}. The Schur deflation scheme \eqref{eq:sd} requires more explanation: matrix Schur deflation is given by \[\bX^{k+1} = \bX^{k} - \bX^{k}\bV_k(\bU_k^T\bX_{k}\bV_k)^{-1}\bU_k^T\bX^k\]
for a target matrix $\bX^k$ and estimated left- and right-singular vectors $\bU_k, \bV_k$ respectively. Na\"ive extension of this formula to the tensor case would require defining the inverse of the quantity $\Xcal^k \ttm{1} \bV_k \ttm{2} \bV_k \ttm{3} \bu_k$; this sort of tensor inverse, however, is not well-defined. Rather than constructing a suitable notion of tensor inverse, we instead apply Schur deflation to each slice of $\Xcal^k$ separately, yielding the Schur deflation scheme given above. 

Using these deflation schemes, we obtain the following general algorithm for the $(r_1, \dots, r_K)$-SS-TPCA problem:
\begin{algorithm}[H]
\begin{itemize}
\item Initialize: $\Xcal^1 = \Xcal$
\item For $k = 1, \dots, K$:
\begin{enumerate}
\item[(i)] Run Algorithm \ref{alg:single_factor_sstpca} on $\Xcal^k$ for rank $r_k$ to obtain ($\bu_k, \bV_k, d_k$)
\item[(ii)] Deflate $\Xcal^k$ using Hotelling \eqref{eq:hd}, projection \eqref{eq:pd}, or Schur \eqref{eq:sd} deflation to obtain $\Xcal^{k+1}$
\end{enumerate}
\item Return $\left\{(\bu_k, \bV_k, d_k)\right\}_{k=1}^K$
\end{itemize}
\caption{Flexible Successive Deflation Algorithm for Multi-Factor $(r_1, \dots, r_K)$-SS-TPCA}
\label{alg:multi_factor_sstpca_general}
\end{algorithm}
The factors estimated by Algorithm \ref{alg:multi_factor_sstpca_general} have the following attractive orthogonality properties:
\begin{appendixtheorem} \label{thm:deflation}
For arbitrary semi-symmetric $\Xcal$, the decomposition estimated by Algorithm \ref{alg:multi_factor_sstpca} satisfies:
\begin{itemize}
    \item Two-way orthogonality, $\langle \Xcal^{k + 1}, \bV_k \circ \bV_k \circ \bu_k \rangle = 0$, at each iteration for all deflation schemes; 
    \item One-way orthogonality for the $\bu$-factor ($\Xcal^{k+1} \ttv{3} \bu_k = \bzero_{p \times p}$) when either projection or Schur deflation is used;  
    \item One-way orthogonality for the $\bV$-factor ($\Xcal^{k+1} \ttm{i} \bV_k = \bzero$ for $i = 1, 2$) when either projection or Schur deflation is used; and
    \item Subsequent orthogonality for the $\bV$-factor ($\Xcal^{k + l + 1} \ttm{i} \bV_k = \bzero$ for $i = 1, 2$ and $l \geq 0$) when Schur deflation is used.
\end{itemize}
The residual tensors $\Xcal^k$ have decreasing norm at each iteration under Hotelling's deflation \eqref{eq:hd} and projection deflation \eqref{eq:pd}: that is, $\|\Xcal^{k+1}\|_F \leq \|\Xcal^k\|_F$ for each $k$. Schur deflation \eqref{eq:sd} also guarantees decreasing norm so long as $\Xcal^k$ remains slicewise positive semi-definite at all $k$.
\end{appendixtheorem} 
The terminology of ``two-way,'' ``one-way,'' and ``subsequent'' orthogonality is due to \citet{Weylandt:2019c}, though we believe this is the first time these techniques have been applied in the tensor context. Proofs of these claims are given in Section \ref{refsec:deflation} of the Supplemental Materials.  We emphasize that these results hold without any assumptions on $\Xcal$ or without any assumptions of the quality of solution identified by Algorithm \ref{alg:single_factor_sstpca} and can be used in other tensor decomposition contexts with minor modification. 

For convenience of the reader, we summarize the key claims of Theorem \ref{thm:deflation} here:
\begin{center} \small
\begin{tabular}{c|ccc}
\toprule
\multirow{2}{*}{\bf Orthogonality}& \multicolumn{3}{c}{\bf Deflation Scheme} \\
 & Hotelling \eqref{eq:hd} & Projection \eqref{eq:pd} & Schur \eqref{eq:sd}\\
\midrule
{\bf Two-Way}: $\langle \Xcal^{k+1}, \bV_k \circ \bV_k \circ \bu_k\rangle = 0$ & \cmark & \cmark & \cmark \\
{\bf One-Way} ($\bu$-factor): $\Xcal^{k+1} \ttv{3} \bu_k = \bzero_{p \times p}$ & \xmark & \cmark & \cmark\\
{\bf One-Way} ($\bV$-factor): $\Xcal^{k+1} \ttm{i} \bV_k = \bzero$ ($i = 1,2$) & \xmark & \cmark & \cmark\\
{\bf Subsequent} ($\bV$-factor): $\Xcal^{k + l + 1} \ttm{i} \bV_k = \bzero$ ($i = 1,2$; $l \geq 0$) & \xmark & \xmark & \cmark\\
\midrule
{\bf True Deflation} $\|\Xcal^{k+1}\|_F \leq \|\Xcal^k\|_F$ & \cmark & \cmark & \cmark\textsuperscript{$\dagger$} \\
\bottomrule
\multicolumn{4}{l}{\footnotesize \textsuperscript{$\dagger$}Under additional conditions that can be checked at run-time.}
\end{tabular}
\end{center}

Proofs of these results are straightforward, if cumbersome, multilinear algebra: the reader is referred to the proofs of these results in Section A.2 of the paper by \citet{Weylandt:2019c} to see the key ideas of each construction in a simpler (matrix) setting. As he notes, \emph{subsequent deflation}, in particular is useful for encouraging approximate orthogonality of greedily estimated factors: if the target of the next SS-TPCA iteration is orthogonal to previous $\bV$, the next estimated $\bV$ terms, which explain that target, are also likely to be nearly orthogonal. Conversely, we have also noted that imposing these additional restrictions may propose unexpected estimation difficulties, most notably decreasing \emph{cumulative percent of variance explained} $\text{CVPE}_k = \|\Xcal^k\|_F^2 / \|\Xcal\|_F^2$: we speculate this may be a useful diagnostic when selecting the number of factors, $K$, but these issues are subtle and best left for future work.

\begin{proof}[Proof of Theorem \ref{thm:deflation}] We first prove two-way orthogonality at each step for Hotelling's deflation: 
\begin{align*}
    \langle \Xcal^{k+1}, \bV_k \circ \bV_k \circ \bu_k \rangle &= \langle \Xcal^{k} - d_k\bV_k \circ \bV_k \circ \bu_k, \bV_k \circ \bV_k \circ \bu_k\rangle \\ 
    &= \langle \Xcal^{k}, \bV_k \circ \bV_k \circ \bu_k\rangle - d_k \langle\bV_k \circ \bV_k \circ \bu_k, \bV_k \circ \bV_k \circ \bu_k\rangle \\
    &= \langle \Xcal^{k}, \bV_k \circ \bV_k \circ \bu_k\rangle - d_k \langle \bV_k \circ \bV_k, \bV_k \circ \bV_k\rangle \langle \bu_k, \bu_k \rangle \\
    &= \langle \Xcal^{k}, \bV_k \circ \bV_k \circ \bu_k\rangle - d_k \|\bV_k \circ \bV_k\|_F^2 \|\bu_k\|^2 \\
    &= \langle \Xcal^{k}, \bV_k \circ \bV_k \circ \bu_k\rangle - d_k * r_k * 1
\end{align*}
Recalling that $d_k = r_k \langle \Xcal^{k}, \bV_k\circ \bV_k \circ \bu_k\rangle$, \emph{cf.} Algorithm \ref{alg:single_factor_sstpca}, this implies $\langle \Xcal^{k+1}, \bV_k \circ \bv_k \circ \bu_k \rangle = 0$ as desired.

Next, we prove one-way orthogonality for the $\bu$ factor under projection deflation:
\begin{align*}
    \Xcal^{k+1} \ttv{3} \bu_k &= \left(\Xcal^{k} \ttm{1} [\bI_{p \times p} - \bV_k \circ \bV_k] \ttm{2} [\bI_{p \times p} - \bV_k \circ \bV_k] \ttm{3} [\bI_{T \times T} - \bu_k\bu_k^T] \right) \ttv{3} \bu_k\\
    &=  \Xcal^{k} \ttm{1} [\bI_{p \times p} - \bV_k \circ \bV_k] \ttm{2} [\bI_{p \times p} - \bV_k \circ \bV_k] \ttv{3} \left([\bI_{T \times T} - \bu_k\bu_k^T]\bu_k\right)\\
    &=  \Xcal^{k} \ttm{1} [\bI_{p \times p} - \bV_k \circ \bV_k] \ttm{2} [\bI_{p \times p} - \bV_k \circ \bV_k] \ttv{3} \left(\bu_k - \bu_k\bu_k^T\bu_k\right)\\
    &= \Xcal^{k} \ttm{1} [\bI_{p \times p} - \bV_k \circ \bV_k] \ttm{2} [\bI_{p \times p} - \bV_k \circ \bV_k] \ttv{3} \bzero_T\\
    &= \bzero_{p \times p}
\end{align*}
where $\bzero_T$ and $\bzero_{p\times p}$ are the all-zero $T$-vector and $p\times p$ matrix respectively. From this, two-way orthogonality is immediate: 
\[\langle \Xcal^{k+1}, \bV_k \circ \bV_k \circ \bu_k\rangle = \langle \Xcal^{k+1} \ttv{3} \bu_k, \bV_k \circ \bV_k\rangle = \langle 0, \bV_k \circ \bV_k \rangle = 0\]

A similar argument gives one-way orthogonality for the $\bV$ factor under projection deflation. Here, we establish orthogonality along the first axis, but the same argument applies for the second axis as well. 
\begin{align*}
    \Xcal^{k+1} \ttm{1} \bV_k &= \left(\Xcal^{k} \ttm{1} [\bI_{p \times p} - \bV_k \circ \bV_k] \ttm{2} [\bI_{p \times p} - \bV_k \circ \bV_k] \ttm{3} [\bI_{T \times T} - \bu_k\bu_k^T] \right) \ttm{1} \bV_k\\
    &= \Xcal^{k} \ttm{1} \left([\bI_{p \times p} - \bV_k\bV_k^T]\bV_k\right)\ttm{2} [\bI_{p \times p} - \bV_k \circ \bV_k] \ttv{3} \bu_k\\
    &= \Xcal^{k} \ttm{1} \left(\bV_k - \bV_k [\bV_k^T\bV_k]\right)\ttm{2} [\bI_{p \times p} - \bV_k \circ \bV_k] \ttv{3} \bu_k\\
    &= \Xcal^{k} \ttm{1} \bzero_{p \times k} \ttm{2} [\bI_{p \times p} - \bV_k \circ \bV_k] \ttv{3} \bzero_T\\
    &= \bzero_{p \times k \times T}
\end{align*}
Recall that we here use a different convention for tensor-matrix multiplication ($\ttm{i}$) than \citet{Kolda:2009} which satisfies the convention $\Xcal \ttm{n} \bA \ttm{n} \bB = \Xcal \ttm{n} (\bA\bB)$ for suitably sized square matrices $\bA, \bB$. The convention used by \citet{Kolda:2009} has $\Xcal \ttm{n} \bA \ttm{n} \bB = \Xcal \ttm{n} (\bB\bA)$; if their convention is used, the second term in the second line of the above would be $\bV_k^T[\bI_{p \times p} - \bV_k\bV_k^T]^T$, which again simplifies to $\bzero$. As before, one-way orthogonality immediately implies two-way orthogonality. 

Finally, we establish the orthogonality properties of Schur deflation: one-way orthogonality for the $\bu$-factor follows essentially the same argument as above: 
\begin{align*}
    \Xcal^{k+1} \ttv{3} \bu_k &= \left(\widetilde{\Xcal}^{k} \ttm{3} [\bI_{T \times T} - \bu_k\bu_k^T] \right) \ttv{3} \bu_k\\
    &= \widetilde{\Xcal}^{k} \ttv{3} \left([\bI_{T \times T} - \bu_k\bu_k^T]\bu_k\right) \\ 
    &= \widetilde{\Xcal}^{k} \ttv{3} \left(\bu_k - \bu_k(\bu_k^T\bu_k)\right) \\ 
    &= \widetilde{\Xcal}^{k} \ttv{3} \bzero_T \\ 
    &= \bzero_{p \times p}
\end{align*}
recalling that $\tilde{\Xcal}^k$ is the tensor formed by Schur deflating each slice of $\Xcal^k$ by $\bV_k$ separately: 
\[\widetilde{\Xcal}^{k+1}_{\cdot\cdot i} = \Xcal^k_{\cdot\cdot i} - \Xcal^k_{\cdot\cdot i}\bV_k(\bV_k^T\Xcal^k_{\cdot\cdot i}\bV_k)^{-1}\bV_k^T \Xcal^k_{\cdot\cdot i} \text{ for all } 1 \leq i \leq T\]
Similarly, for the $\bV_k$ factor, we note that 
\begin{align*}
    \Xcal^{k+1} \ttm{1} \bV_k &= \left(\widetilde{\Xcal}^{k} \ttm{3} [\bI_{T \times T} - \bu_k\bu_k^T] \right) \ttm{1} \bV_k\\
    \Xcal^{k+1} \ttm{1} \bV_k &= \underbrace{\left(\widetilde{\Xcal}^{k} \ttm{1} \bV_k\right)}_{=\overline{\Xcal^k}} \ttm{3} [\bI_{T \times T} - \bu_k\bu_k^T]
\end{align*}
Here, each slice of $\overline{\Xcal}^k$ is given by 
\begin{align*}
    \overline{\Xcal}^k_{\cdot\cdot i} &= \bV_k^T\Xcal^k_{\cdot\cdot i} - \bV_k^T\Xcal^k_{\cdot\cdot i}\bV_k(\bV_k^T\Xcal^k_{\cdot\cdot i}\bV_k)^{-1}\bV_k^T \Xcal^k_{\cdot\cdot i} \\
    &= \bV_k^T\Xcal^k_{\cdot\cdot i} - \bV_k^T\Xcal^k_{\cdot\cdot i} \\
    &= \bzero_{k \times p}
\end{align*}
implying $\overline{\Xcal}^k = \bzero_{k \times p \times T}$ and hence $\Xcal^{k+1} \ttm{1} \bV_k = \bzero_{k \times p \times T}$. The same argument holds along the second axis of $\Xcal$. 

We can take this as a base case to inductively establish subsequent orthogonality for the $\bV$ factor: assuming $\Xcal^{k+l} \ttm{1} \bV_{k} = \bzero$, we consider $\Xcal^{k + l + 1} \ttm{1} \bV_k$:
\begin{align*}
    \Xcal^{k + l + 1} \ttm{1} \bV_k &= \underbrace{\widetilde{\Xcal}^{k + l + 1} \ttm{1} \bV_k}_{=\hat{\Xcal}^{k + l + 1}} \ttm{3} [\bI_{T \times T} - \bu_k\bu_k^T]
\end{align*}
where $\hat{\Xcal}^{k + l + 1}$ is a semi-symmetric tensor, each slice of which is given by
\[\hat{\Xcal}^{k + l + 1}_{\cdot\cdot i} = \bV_k^T\Xcal^{k + l}_{\cdot\cdot i} - \bV_k^T\Xcal^{k + l}_{\cdot\cdot i}\bV_k(\bV_k^T\Xcal^{k + l}_{\cdot\cdot i}\bV_k)^{-1}\bV_k^T\Xcal^{k + l}_{\cdot\cdot i}\]
Under the inductive hypothesis, we note that \[\bV_k^T\Xcal^{k + l}_{\cdot \cdot i} = \left[\Xcal^{k + l} \ttm{1} \bV_k\right]_{\cdot\cdot i} = \bzero_{\cdot\cdot i} = \bzero_{k \times p}\]
and hence $\hat{\Xcal}^{k + l + 1} = \bzero$ so $\Xcal^{k + l + 1} \ttm{1} \bV_k = \bzero_{k \times p \times T}$ for all $l \geq 0$.

We first show that Hotelling's deflation gives a true deflation directly: 
\begin{align*}
    \|\Xcal^{k}\|_F &\geq \|\Xcal^{k+1}\|_F \\ 
    \|\Xcal^k\|_F^2 - \|\Xcal^k - d_k \bV_k \circ \bV_k \circ \bu_k\|_F^2 & \geq 0\\
    \|\Xcal^k\|_F^2 - \left(\|\Xcal^k\|_F^2 - 2d_k \langle \Xcal^k, \bV_k \circ \bV_k \circ \bu_k\rangle + d_k^2 \|\bV_k \circ \bV_k \circ \bu_k\|_F^2\right) & \geq 0\\
    2\langle \Xcal^k, \bV_k \circ \bV_k \circ \bu_k\rangle &\geq d_k \|\bV_k \circ \bV_k \circ \bu_k\|_F^2 \\ 
    2\langle \Xcal^k, \bV_k \circ \bV_k \circ \bu_k\rangle &\geq d_k r_k
\end{align*}
Recalling that $d_k = r_k^{-1} \langle \Xcal^k, \bV_k \circ \bV_k \circ \bu_k\rangle$, the desired result follows immediately from the fact that $\langle \Xcal^k, \bV_k \circ \bV_k \circ \bu_k\rangle \geq 0$.\footnote{This can always be ensured by  replacing $\bu_k$ with $-\bu_k$ as needed.} 

For projection and Schur deflation, a different analysis is needed: we first note that 
$\|\bA\|_F \geq \|\bA(\bI - \bu\bu^T)\|_F$ for any real matrix $\bA$ and any unit vector $\bu$.\footnote{Note that $\|\bA\|_F^2 = \Tr(\bA^T\bA) = \sum_{i=1}^n \sigma_i(\bA)^2$ while $\|\bA(\bI - \bu\bu^T)\|_F^2 = \Tr(\bA^T\bA) - \Tr(\bu^T\bA^T\bA\bu)$. This last term is minimized when $\bu$ is the trailing eigenvector of $\bA^T\bA$, so $\|\bA(\bI - \bu\bu^T)\|_F^2 \leq \sum_{i=1}^{n-1} \sigma_i(\bA)^2$, clearly giving the desired result. Here we use the convention $\sigma_1(\bB) \geq \sigma_2(\bB) \geq \dots \geq \sigma_n(\bB) \geq 0$ to order the singular values of a matrix.} Applying this result separately to each slice of a tensor and recalling that the Frobenius norm adds in quadrature across the slices, we have $\|\Xcal \ttm{3} (\bI - \bu\bu^T)\|_F \leq \|\Xcal\|_F$ for any $\Xcal$ and unit vector $\bu$. For projection deflation, we note that \[\left\|\left\{\Xcal^k \ttm{1} (\bI - \bV_k \circ \bV_k)\ttm{2} (\bI - \bV_k \circ \bV_k)\right\}_{\cdot\cdot i}\right\|_F < \|\Xcal^k_{\cdot\cdot i}\|_F\]
for each slice separately by essentially the same argument as above. Hence, we have
\[\|\Xcal^k\|_F \geq \|\Xcal^k \ttm{1} [\bI - \bV_k\circ \bV_k] \ttm{2} [\bI - \bV_k \circ \bV_k]\| \geq \|\Xcal^k \ttm{1} [\bI - \bV_k\circ \bV_k] \ttm{2} [\bI - \bV_k \circ \bV_k] \ttm{3} [\bI - \bu_k \bu_k]\| \geq \|\Xcal^{k+1}\|\]
as desired. For Schur deflation, we note that  $\|\bA\| > \|\bA - \bA \bV(\bV^T\bA\bV)^{-1}\bV^T\bA\|$ for any positive semi-definite matrix $\bA$ and orthogonal matrix $\bV$ and apply this slicewise in a similar fashion. \qedhere
\end{proof} 
We note that the assumption of slicewise positive semi-definitness is only used in the proof of decreasing norm and is not required for the orthogonality results. 

Finally, we note that the single-factor rank-$r$ SS-TPCA is statistically identifiable, with paramterization $(\bu, \bV_*, \sigma) \in \overline{\B}^T \times \mathscr{G}(r, \R^p) \times \R_{> 0}$, where $\mathscr{G}(r, \R^p)$ is the Grassmanian manifold of all $r$-dimensional sub-spaces of $\R^p$ or equivalently the quotient manifold $\Vcal^{p \times r} / \Ocal(r)$, where $\Ocal(r)$ is the set of $r \times r$ orthogonal matrices. Identifiability conditions for the multi-rank SS-TPCA are more subtle and beyond the scope of this work, but we expect that considerations for multi-factor identifiability would rely, in part, on the literature discussing uniqueness of tensor decompositions; see Sections 3.2 and Sections 4.3 of the review by \citet{Kolda:2009} for an introduction to these issues.

\section{Elementary Finite-Sample Results for Classical SVD-PCA} \label{app:pca}
In this section, we apply the celebrated $\sin \Theta$ theorem of \citet{Davis:1970} to analyze the SVD approach to PCA under \textsc{iid} sub-Gaussian noise. We use the formulation of \citet{Yu:2015} to simplify the assumptions imposed on both the underlying signal matrix and the noisy observation thereof. 

With these tools, we establish the following consistency result for SVD-based PCA:
\begin{appendixtheorem} \label{thm:pca}
Let $\bX_* \in \R^{n \times p}$ admit an exact rank-one decomposition as $\bX_* = d \bu_*\bv^T_*$ where $d \in \R_{> 0}$ is a measure of signal strength and $\bu_* \in \B^n$ and $\bv_* \in \B^p$ are unit vectors of length $n$ and $p$ respectively. Suppose $\bX = \bX_* + \bE$ be observed, where the elements of $\bE$ are independent mean-zero $\sigma^2$-sub-Gaussian random variables. If $\hat{\bu}, \hat{\bv}$ are the leading left- and right-singular vectors of $\bX$ respectively, then, with probability at least $1 - e^{-t^2}$, we have
\begin{align*}
    \min_{\epsilon = \pm 1} \frac{\|\bu_* - \epsilon \hat{\bu}\|}{\sqrt{n}} &\lesssim \max\left\{\sigma\frac{1 + \sqrt{p} + t}{d}, \sigma^2 \frac{(\sqrt{n} + \sqrt{p} + t)^2}{d^2\sqrt{n}}\right\} \\ 
    \min_{\epsilon = \pm 1} \frac{\|\bv_* - \epsilon \hat{\bv}\|}{\sqrt{p}} &\lesssim \max\left\{\sigma\frac{1 + \sqrt{n} + t}{d}, \sigma^2 \frac{(\sqrt{n} + \sqrt{p} + t)^2}{d^2\sqrt{p}}\right\}
\end{align*}
where $\lesssim$ denotes an inequality holding up to a universal constant factor.
\end{appendixtheorem}
In the case with signal-to-noise ratio greater than $1$, \emph{i.e.}, $d \gg \sigma(\sqrt{n} + \sqrt{p})$, the left hand branch dominates and we obtain the following limiting behavior: with probability at least $1 - e^{-t^2}$
\begin{equation}
  \begin{split}
    \min_{\epsilon = \pm 1} \frac{\|\bu_* - \epsilon \hat{\bu}\|}{\sqrt{n}} &\lesssim \sigma\frac{\sqrt{p} + t}{d} \\ 
    \min_{\epsilon = \pm 1} \frac{\|\bv_* - \epsilon \hat{\bv}\|}{\sqrt{p}} &\lesssim \sigma \frac{\sqrt{n} + t}{d}.
  \end{split} \label{eqn:pca_svd}
\end{equation}
We note that this is essentially the same result as SS-TPCA given in Theorems \ref{thm:consistency} and \ref{thm:vconsistency}.

The proof of Theorem \ref{thm:pca} builds upon the following proposition, which is a simple special case of Theorem 4 of \citet{Yu:2015}, itself a restatement of \citegenitive{Wedin:1972} celebrated result.
\begin{appendixproposition} \label{prop:daviskahan_svd}
Let $\bX_* \in \R^{n \times p}$ admit an exact rank-one decomposition as $\bX_* = d \bu_*\bv^T_*$ where $d \in \R_{> 0}$ is a measure of signal strength and $\bu_* \in \B^n$ and $\bv_* \in \B^p$ are unit vectors of length $n$ and $p$ respectively. Suppose $\bX = \bX_* + \bE$ be observed, for some fixed but unknown $\bE$. If $\hat{\bu}, \hat{\bv}$ are the leading left- and right-singular vectors of $\bX$ respectively, then,
\begin{align*}
    \min_{\epsilon = \pm 1} \|\bu_* - \epsilon \hat{\bu}\| &\leq 2^{5/2} \frac{\opnorm{\bE}}{d} + 2^{3/2} \frac{\opnorm{\bE}^2}{d^2} \\ 
    \min_{\epsilon = \pm 1} \|\bv_* - \epsilon \hat{\bv}\| &\leq 2^{5/2} \frac{\opnorm{\bE}}{d} + 2^{3/2} \frac{\opnorm{\bE}^2}{d^2}
\end{align*}
\end{appendixproposition}
\begin{proof}
This result follows from  the second claim of Theorem 4 of \citet{Yu:2015}, where we take $r = s = 1$, $d = 1$, $\sigma_1 = \sigma_r = \sigma_s = d$, $\sigma_2 = \sigma_{r - 1} = 0$, and $\sigma_{s + 1} = \infty$ and noting $d^2 < \sigma_{s + 1}^2 - d^2$ in the denominator. We take $\min(\opnorm{\bE}, \|\bE\|_F) \leq \opnorm{\bE}$ in the numerator. \qedhere
\end{proof}
With this in hand, we are ready to prove Theorem \ref{thm:pca}:
\begin{proof}[Proof of Theorem \ref{thm:pca}]
We begin by controlling $\opnorm{\bE}$ under the assumptions that each element of $\bE$ is independently mean-zero $\sigma^2$-sub-Gaussian. Applying Theorem 4.4.5 of \citet{Vershynin:2018}\footnote{See also Example 5.19 of \citet{Wainwright:2019}.}, we have
\[\opnorm{\bE} \lesssim \sigma \left(\sqrt{n} + \sqrt{p} + t\right)\]
with probability at least $1 - e^{-t^2}$. Note that here, each element of $\bE$ is $\sigma^2$-sub-Gaussian so $\|E_{ij}\|_{\psi_2} = \sigma$ for all $i, j$ and hence $K = \sigma$ in \citeauthor{Vershynin:2018}'s result. Combining this bound with Proposition \ref{prop:daviskahan_svd}, we have
\begin{align*}
    \min_{\epsilon = \pm 1} \|\bu_* - \epsilon \hat{\bu}\| &\lesssim 2^{5/2}\sigma \frac{\sqrt{n} + \sqrt{p} + t}{d} + 2^{3/2}\sigma^2 \frac{(\sqrt{n} + \sqrt{p} + t)^2}{d^2} \\ 
    \min_{\epsilon = \pm 1} \|\bv_* - \epsilon \hat{\bv}\| &\lesssim 2^{5/2}\sigma \frac{\sqrt{n} + \sqrt{p} + t}{d} + 2^{3/2}\sigma^2 \frac{(\sqrt{n} + \sqrt{p} + t)^2}{d^2}
\end{align*}
with probability as least $1 - e^{-t^2}$. Simplifying gives
\begin{align*}
    \min_{\epsilon = \pm 1} \|\bu_* - \epsilon \hat{\bu}\| &\lesssim 8\max\left\{\sigma\frac{\sqrt{n} + \sqrt{p} + t}{d},  \sigma^2\frac{(\sqrt{n} + \sqrt{p} + t)^2}{d^2}\right\}  \\ 
    \min_{\epsilon = \pm 1} \|\bv_* - \epsilon \hat{\bv}\| &\lesssim 8\max\left\{\sigma\frac{\sqrt{n} + \sqrt{p} + t}{d},  \sigma^2\frac{(\sqrt{n} + \sqrt{p} + t)^2}{d^2}\right\} 
\end{align*}
Finally, if we normalize each side to root mean squared error, we obtain the desired results: 
\begin{align*}
    \min_{\epsilon = \pm 1} \frac{\|\bu_* - \epsilon \hat{\bu}\|}{\sqrt{n}} &\lesssim 8\max\left\{\sigma\frac{1 + \sqrt{p} + t}{d}, \sigma^2 \frac{(\sqrt{n} + \sqrt{p} + t)^2}{d^2\sqrt{n}}\right\} \\ 
    \min_{\epsilon = \pm 1} \frac{\|\bv_* - \epsilon \hat{\bv}\|}{\sqrt{p}} &\lesssim 8\max\left\{\sigma\frac{1 + \sqrt{n} + t}{d}, \sigma^2 \frac{(\sqrt{n} + \sqrt{p} + t)^2}{d^2\sqrt{p}}\right\} \qedhere
\end{align*}
\end{proof}

\subsection{Comparison with Covariance PCA}
The implications of Equation \eqref{eqn:pca_svd} above may be somewhat counterintuitive for readers more familiar with the covariance (eigendecomposition) approach to PCA. Indeed, in that setting, having more samples, $n$, improves the estimation of the $p$-vector $v$. This disparity comes from subtle distinctions in the low-rank mean model considered here, where the signal is fixed, and the standard spiked covariance model, where the signal becomes more apparent with more samples. 

To see this, note that under the assumptions of Theorem \ref{thm:pca}, the signal term $\bX_*$ has fixed strength while the size of the noise grows as $\sqrt{n} + \sqrt{p}$, yielding an effective signal to noise of $\text{SNR} = d / (\sqrt{n} + \sqrt{p})$.\footnote{Note that we fix the marginal variance of each element of $\bE$ rather than scaling it with $n$ or $p$.} By contrast, under the spiked covariance model, with $\bX$ having $n$ rows each of which are independently sampled from a normal distribution with mean $\bzero$ and covariance $\sigma^2 \bI_{p \times p} + \theta \bv\bv_*$, we have \[\Cov[\bX^T\bX/n] = \Cov\left[\frac{1}{n}\sum_{i=1}^n \bx_i\bx_i^T\right] = \frac{1}{n}^2 \sum_{i=1}^n \sigma^2 \bI_{p \times p} + \theta \bv\bv_* = \frac{\sigma^2 \bI_{p \times p} + \theta\bv\bv_*}{n}\]
yielding a signal to noise ratio \emph{decreasing} rather than increasing in $n$. For a more refined analysis, see Theorem 6.5 of \citet{Wainwright:2019} and the surrounding discussion.

While it is unnecessary to analyze an eigendecomposition approach to Theorem \ref{thm:pca} because the SVD and eigendecomposition approaches give equivalent answers, we outline such an approach, showing that it leads to \emph{worse} results than those obtained via the SVD. Suppose we wish to establish a result like the following:
\begin{appendixtheorem}[Informal analysis of the covariance estimator under the low-rank mean model.] \label{thm:pca_eigen}
Let $\bX_* \in \R^{n \times p}$ admit an exact rank-one decomposition as $\bX_* = d \bu_*\bv^T_*$ where $d \in \R_{> 0}$ is a measure of signal strength and $\bu_* \in \B^n$ and $\bv_* \in \B^p$ are unit vectors of length $n$ and $p$ respectively. Suppose $\bX = \bX_* + \bE$ be observed, where the elements of $\bE$ are independent and $\sigma^2$-sub-Gaussian. If $\hat{\bv}$ is the leading eigenvector of $\bX^T\bX$, then, with probability at least $1 - e^{-t^2}$, we have
\begin{align*}
    \min_{\epsilon = \pm 1} \frac{\|\bu_* - \epsilon \hat{\bu}\|}{\sqrt{n}} &\lesssim \dots \\ 
    \min_{\epsilon = \pm 1} \frac{\|\bv_* - \epsilon \hat{\bv}\|}{\sqrt{p}} &\lesssim \dots
\end{align*}
\end{appendixtheorem}

As before, we use a variant of the Davis-Kahan theorem to analyze this model: 
\begin{appendixproposition} \label{prop:daviskahan_eigen}
Let $\bX_* \in \R^{n \times p}$ admit an exact rank-one decomposition as $\bX_* = d \bu_*\bv^T_*$ where $d \in \R_{> 0}$ is a measure of signal strength and $\bu_* \in \B^n$ and $\bv_* \in \B^p$ are unit vectors of length $n$ and $p$ respectively and $n > p$. Suppose $\bX = \bX_* + \bE$ be observed for some fixed, but unobserved, $\bE$. If $\hat{\bv}$ is the leading eigenvector of $\bX^T\bX$, then 
\[\min_{\epsilon = \pm 1} \|\bv_* - \epsilon \hat{\bv}\| \leq \frac{2^{3/2} \opnorm{\bE^T\bX_* + \bX_*^T\bE + \bE^T\bE}}{d^2}\]
\end{appendixproposition}
\begin{proof} This proof follows immediately from Theorem 2 of \citet{Yu:2015} using the same notation substitutions as in the proof of Proposition \ref{prop:daviskahan_svd}, with the exception of noting that $\lambda_{\max}(\bX_*^T\bX_*) = \sigma_{\max}(\bX_*)^2 = d^2$. \qedhere
\end{proof}
We are now ready to analyze Theorem \ref{thm:pca_eigen}: as before, the bulk of the proof is establishing appropriate high-probability bounds on the noise terms.
\begin{proof}[Analysis of Quasi-Theorem \ref{thm:pca_eigen}]
Applying Proposition \ref{prop:daviskahan_eigen} to our estimator of interest, we need to bound $\opnorm{\bE^T\bX_* + \bX_*^T\bE + \bE^T\bE}$. Using a triangle inequality, we bound the linear terms and the quadratic term separately: 
\[\opnorm{\bE^T\bX_* + \bX_*^T\bE + \bE^T\bE} \leq 2\opnorm{\bX_*^T\bE} + \opnorm{\bE^T\bE}\]
The linear term is relatively straightforward: we note that \[\opnorm{\bX_*^T\bE} = \opnorm{d\bu\bv^T\bE} \leq d \|\bu\| \|\bV\| \opnorm{\bE} = d\opnorm{\bE}\] and, as before, we have a probabilistic bound on $\opnorm{\bE}$ leading to $\opnorm{\bX_*^T\bE} \lesssim \sigma(\sqrt{n} + \sqrt{p} + t)$ with probability at least $1 - e^{-t^2}$, which simplifies to \[P\left\{\opnorm{\bX_*^T\bE} \gtrsim d\sigma(\sqrt{n} + t)\right\} \leq 1 - e^{-4t^2}\] in our case.

The quadratic term is more difficult: we note that $\opnorm{\bE^T\bE}$ will be a sub-exponential random variable and hence concentrates as 
\[P\left[\opnorm{\bE^T\bE - \sigma^2 \bI} \lesssim \sqrt{\frac{p}{n}} + \delta \right] \geq 1 - ce^{-n\min{\delta, \delta^2}}\]
for some absolute $c$. This follows from Theorem 6.5 of \citet{Wainwright:2019}, absorbing his $c_3$ into the definition of $\delta$ and his $c_1$ into the hidden constant term and noting that $\sqrt{p / n} > p / n$ by assumption. Applying the (reverse) triangle inequality to the right hand side, we find
\[P\left\{\opnorm{\bE^T\bE} \gtrsim \sigma^2(1 + \sqrt{\frac{p}{n}} + \delta)\right\} 1 - ce^{-n\min{\delta, \delta^2}}\]
Heuristically, this bound makes sense: the constant term ($\sigma^2)$ captures the expected operator norm of $\bE^T\bE$ while the $\sqrt{p / n}$ term captures expected fluctuations (decreasing as $n \to \infty$) and the $\delta$ term and form of the high-probability bound are as expected for squared sub-Gaussian (sub-exponential) quantities. \qedhere
\end{proof}
We do not provide precise results here, but it is clear that the error will be dominated by the sub-exponential $\bE^T\bE$ term, which concentrates less efficiently than the sub-Gaussian $\bE$ term of the low-rank model. Interestingly, this finding has parallels in the numerical analysis literature, where computing $\bX^T\bX$ and taking its eigendecomposition rather than taking a singular value decomposition of $\bX$ directly has the effect of squaring the problem's condition number and decreasing numerical accuracy.

\section{Proofs for Section \ref{sec:theory} - \nameref{sec:theory}} \label{refsec:proofs}
In this section, we present the complete proofs of Theorems \ref{thm:consistency} and \ref{thm:vconsistency}, giving more detail than in the main text above. Before proceeding, we state for completeness the variant of the Davis-Kahan \citeyearpar{Davis:1970} theorem  due to \citet{Yu:2015}, specialized for our case:
\begin{restatable}{appendixproposition}{daviskahan} \label{prop:daviskahan}
Suppose $\bSigma^* \in \R^{p \times p}_{\succeq 0}$ has rank $r < p$ and has all non-zero eigenvalues at least $\lambda > 0$. Suppose further that $\hat{\bSigma} = \bSigma^* + \bE$ is an estimate of $\bSigma^*$. Then 
\[\|\sin \Theta(\bV^*, \hat{\bV})\|_F \leq \frac{2 r^{1/2} \opnorm{\bE}}{\lambda}\]
where $\bV^*, \hat{\bV}$ are matrices composed of the $r$ leading eigenvectors of $\bV^*, \hat{\bV}$ respectively and $\Theta (\bV^*, \hat{\bV})$ denotes the $r \times r$ diagonal matrix whose entries are the principal angles between the spaces spanned $\bV^*$ and $\hat{\bV}$ and where $\sin(\cdot)$ is applied elementwise. Furthermore, 
there exists an orthogonal matrix $\bO \in \Vcal^{r \times r}$ such that
\[\|\bV^* - \hat{\bV}\bO\|_F \leq \frac{2^{3/2}r^{1/2}\opnorm{\bE}}{\lambda}.\]
\end{restatable}
\begin{proof} This proof follows immediately from Theorem 1 of \citet{Yu:2015}, with $r, s, d$ in their notation being equal to $1, r, r$ in our notation. The denominator of their bound is \[\min(\lambda_0(\bSigma^*) - \lambda_1(\bSigma^*), \lambda_r(\bSigma^*) - \lambda_{r + 1}(\bSigma^*)) \geq \min(\infty - \lambda, \lambda - 0) = \lambda,\] which gives the denominator in our statement above. Finally, we use only the operator norm bound on $\bE$ as it is easier to analyze under our noise model, but the Frobenius norm could be used as well. \qedhere 
\end{proof}
With this result in hand, we are able to prove Theorem \ref{thm:consistency}. To simplify the proof, we first establish the following non-stochastic bound, which extends Proposition \ref{prop:deterministicconsistency_u} in the main text to also provide bounds on $\hat{\bV}$:
\begin{appendixproposition} \label{appprop:consistency} Suppose $\Xcal = d\, \bV_* \circ \bV_* \circ \bu_* + \Ecal$ for a unit-norm $T$-vector $\bu_*$, a $p \times r$ orthogonal matrix $\bV_*$ satisfying $\bV_*^T\bV_* = \bI_{r \times r}$, $r \in \R_{\geq 0}$, and $\Ecal \in \R^{p \times p \times T}$ a semi-symmetric tensor. Then the result of Algorithm \ref{alg:single_factor_sstpca} applied to $\Xcal$ satisfies the following: 
\begin{align*}
  \min_{\epsilon = \pm 1} \|\epsilon \bu^* - \hat{\bu}\|_2 &\leq \frac{8\sqrt{2}\ropnorm{\Ecal}}{d(1 - c)} \\
  \|\bV^*\bO - \hat{\bV}\|_F &\leq \frac{8\sqrt{2r}\ropnorm{\Ecal}}{d(1 - c)}
\end{align*}
so long as $\ropnorm{\Ecal} < d$ and 
\[|1 - \langle \bu^{(0)}, \bu_*\rangle | \leq \tan^{-1}(0.5) - \frac{5 \ropnorm{\Ecal}}{d(1 - c)}\] for some arbitrary $c < 1$.
\end{appendixproposition}

The structure of our proof is straightforward: we use the Davis-Kahan theorem to analyze a single iteration of the $\bV$ and $\bu$-update steps of Algorithm \ref{alg:single_factor_sstpca}, showing that the error at each step can be bounded by a sum of a term depending on the error from the previous iteration and a term depending on $\ropnorm{\Ecal}/d$. Under some conditions which can be interpreted as being in a ``basin of attraction,'' we show that the iterates contract to a small ball around $\bV_*$ and $\bu_*$. Finally, we show that the initialization condition on $\bu^{(0)}$ and some assumptions on the relative magnitude of $\ropnorm{\Ecal}$ and $d$ suffice to ensure that all iterates remain in the basin of attraction. Combining these steps, we obtain a proof of Proposition \ref{appprop:consistency}

\begin{proof}  For simplicity, and with minimal loss of generality, we specialize our proof to the rank-1 case. Comments on the general rank-$r$ case appear at the end of each section of the proof. Our proof proceeds in four parts: I. Analysis of the $\bV$-Update; II. Analysis of the $\bu$-Update; III. Iterative Error Analysis; and IV. Condition Checking.

\noindent\textbf{I. Analysis of $\bV$-Update.} Recall that, in the rank-1 case, the $\bv$-update is given by 
\[\bv^{(k+1)} = \textsf{eigen}(\Xcal \ttv{3} \bu^{(k)})\] 
where $\textsf{eigen}(\cdot)$ denotes the leading eigenvector. Under our model, this can be simplified as
\[\Xcal \ttv{3} \bu^{(k)} = (d \langle \bu_*, \bu^{(k)}\rangle) \bv_* \circ \bv_* + \Ecal \ttv{3} \bu^{(k)}\] We apply Proposition \ref{prop:daviskahan} with $\bSigma^* = d\, \bv_* \circ \bv_*$ and $\hat{\bSigma} = \Xcal \ttv{3} \bu^{(k)}$. This implies $\bE = d(\langle \bu^{(k)}, \bu_*\rangle - 1)\bv_* \circ \bv_* + \Ecal \ttv{3} \bu^{(k)}$ and hence
\begin{align*}
    \opnorm{\bE} &= \left\|d(\langle \bu^{(k)}, \bu_*\rangle - 1)\bv_* \circ \bv_* + \Ecal \ttv{3} \bu^{(k)}\right\|_{\text{op}} \\
    &\leq |d| \, \left|\langle \bu^{(k)}, \bu_*\rangle - 1\right| \left\|\bv^* \circ \bv^*\right\|_{\text{op}} + \|\Ecal \ttv{3} \bu^{(k)}\|_{\text{op}} \\
    &\leq d \left|\langle \bu^{(k)}, \bu_*\rangle - 1\right| + \opnorm{\Ecal}
\end{align*}
Hence Proposition \ref{prop:daviskahan} implies 
\[|\sin \angle(\bv^*, \bv^{(k+1)})| \leq 2\left|\langle \bu^{(k)}, \bu_*\rangle - 1\right| + \frac{2\opnorm{\Ecal}}{d} \]

The extension to the the rank-$r$ case is straightforward: 
\[\left\|\sin \Theta(\bv^*, \bv^{(k+1)})\right\|_F \leq 2\sqrt{r}\left|\langle \bu^{(k)}, \bu_*\rangle - 1\right| + \frac{2\sqrt{r}\ropnorm{\Ecal}}{d} \] The only noteworthy change is an additional $\sqrt{r}$-term arising in the final Davis-Kahan bound. 

\noindent\textbf{II. Analysis of $\bu$-Update.} Recall that the $\bu$-update is given by \[\bu^{(k+1)} = \textsf{Norm}([\Xcal; \bV^{(k+1)}])\] In the rank-1 case, this can be rewritten and simplified as
\begin{align*}
  \bu^{(k+1)} &= \textsf{Norm}([\Xcal; \bV^{(k+1)}]) \\
  &= \textsf{Norm}([d\, \bv_* \circ \bv_* \circ \bu_* + \Ecal; \bv^{(k+1)}]) \\
  &= \textsf{Norm}(d [\bv_* \circ \bv_* \circ \bu_*; \bv^{(k+1)}] + [\Ecal; \bv^{(k+1)}]) \\
  &= \textsf{Norm}(d\, \bv_* \circ \bv_* \circ \bu_* \ttv{1} \bv^{(k+1)} \ttv{2} \bv^{(k+1)} + \Ecal \ttv{1} \bv^{(k+1)} \ttv{2} \bv^{(k+1)}) \\
  &= \textsf{Norm}\left(\underbrace{d \bu_* |\langle \bv_*, \bv^{(k+1)} \rangle|^2 + \Ecal \ttv{1} \bv^{(k+1)} \ttv{2} \bv^{(k+1)}}_{=\tilde{\bu}^{(k+1)}}\right)
\end{align*}
This quantity is difficult to analyze directly due to the non-linear normalization step, but we can circumvent these difficulties by a ``reverse'' use of the Davis-Kahan theorem. By construction, $\bu^{(k+1)}$ is the leading eigenvector of $\tilde{\bu}^{(k+1)} \circ \tilde{\bu}^{(k+1)}$ and $\bu_*$ is the leading eigenvector of $d^2 \bu_* \circ \bu_*$. Note that
\[\tilde{\bu}^{(k+1)} \circ \tilde{\bu}^{(k+1)} = (d')^2 \bu_* \circ \bu_* + d' \bu_* \circ \be + d' \be \circ \bu^* + \be \circ \be\]
where $d' = d|\langle  \bv_*, \bv^{(k+1)}\rangle|^2$ and $\be = \Ecal \ttv{1} \bv^{(k+1)} \ttv{2} \bv^{(k+1)}$. Plugging these into Proposition \ref{prop:daviskahan},\footnote{Note that we can rescale the matrices $\tilde{\bu}^{(k+1)} \tilde{\bu}^{(k+1)}$ and $\bu_* \circ \bu_*$ however we wish before applying the Davis-Kahan theorem because scalar descaling does not change the eigenvectors of each matrix. The scale factor chosen here, $d'$, minimizes the error given by the Davis-Kahan theorem while remaining algebraically tractable.} we have
\begin{align*}
    \opnorm{\bE} &= \left\|d^2 \bu_* \circ \bu_* - \tilde{\bu}^{(k+1)} \circ \tilde{\bu}^{(k+1)}\right\| \\
    &= \left\|d^2(1 - \langle \bv_*, \bv^{(k+1)}\rangle^4)\bu_*\circ \bu_* + d' \bu_* \circ \be + d' \be \circ \bu^* + \be \circ \be\right\|_{\text{op}} \\
    &\leq d^2|(1 - \langle \bv_*, \bv^{(k+1)}\rangle^4)|\|\bu_* \circ \bu_*\|_{\text{op}} + 2d' \|\bu_* \circ \be\|_{\text{op}} + \|\be\circ\be\|_{\text{op}} \\
    &\leq d^2|(1 - \langle \bv_*, \bv^{(k+1)}\rangle^4)|\|\bu_*\|^2 + 2d' \|\bu_*\| \|\be\| + \|\be\|^2
\end{align*}
which implies
\begin{align*}|\sin \angle(\bu_*, \bu^{(k+1)})| &\leq \frac{2 \opnorm{\bE}}{d^2} \\
&\leq \frac{2\left[d^2|(1 - \langle \bv_*, \bv^{(k+1)}\rangle^4)|\|\bu_*\|^2 + 2d' \|\bu_*\| \|\be\| + \|\be\|^2\right]}{d^2}\\
&= 2|(1 - \langle \bv_*, \bv^{(k+1)}\rangle^4)| + \frac{4d' \|\be\|}{d^2} + \frac{2\|\be\|^2}{d^2} \\
&\leq 2|(1 - \langle \bv_*, \bv^{(k+1)}\rangle ^4)| + \frac{4 \opnorm{\Ecal}}{d} + \frac{2\opnorm{\Ecal}^2}{d^2}
\end{align*}
where the final inequality follows from the fact that $d' = d(1 - \langle \bv^*, \bv^{(k+1)}\rangle^2) < d$ and $\|\be\| = \|\Ecal \ttv{1} \bv^{(k+1)} \ttv{2} \bv^{(k+1)}\| \leq \opnorm{\Ecal}$. 

This is the step of the proof requiring the most cumbersome algebra, and hence the hardest to extend to the rank-$r$ case. The major difficulty is in substituting the $[\Xcal, \bV^{(k+1)}]$ for the simpler $\Xcal \ttv{1} \bv^{(k+1)} \ttv{2} \bv^{(k+1)}$ construction, but both are linear in the $\Xcal$ argument and the same argument essentially holds, with the normalized Frobenius inner product on matrices $\langle \bA, \bB \rangle = \Tr(\bA^T\bB) / r$ taking the place of the Euclidean inner product. This gives the error bound: 
\[|\sin \angle(\bu_*, \bu^{(k+1)})| \leq 2(1 - \langle \bV_*, \bV^{(k+1)}\rangle^4) + \frac{4r\opnorm{\Ecal}}{d} + \frac{2r^2\ropnorm{\Ecal}^2}{d^2}\]
where the $\ropnorm{\Ecal}$ terms are scaled by $r$ to account for the fact that we are now using $r$ vectors instead of just one. Note that the Davis-Kahan bound does not pick up an extra $\sqrt{r}$ since the $\bu$ factor remains univariate even in the rank-$r$ decomposition.

\noindent\textbf{III. Iterative Error Analysis.} The previous two parts gave us the following accuracy bounds: 
\begin{align*}
    |\sin \angle(\bv^*, \bv^{(k+1)})| &\leq 2\left|\langle \bu^{(k)}, \bu_*\rangle - 1\right| + \frac{2\opnorm{\Ecal}}{d}  \\
    |\sin \angle(\bu_*, \bu^{(k+1)})| &\leq 2(1 - \langle \bv_*, \bv^{(k+1)}\rangle ^4) + \frac{4 \opnorm{\Ecal}}{d} + \frac{2\opnorm{\Ecal}^2}{d^2}
\end{align*}
or equivalently: 
\begin{align}
    |\sin \theta_{\bv_{k+1}}| &\leq 2|1 - \cos \theta_{\bu_{k}}| + \frac{2\opnorm{\Ecal}}{d} \label{eqn:sin_v}\\
    |\sin \theta_{\bu_{k+1}}| &\leq 2|1 - \cos^4\theta_{\bv_{k+1}}| + \frac{4\opnorm{\Ecal}}{d} + \frac{2\opnorm{\Ecal}^2}{d^2} \label{eqn:sin_u}
\end{align}
where $\theta_{\bv_k} = \angle(\bv^{(k)}, \bv_*) = \cos^{-1}\langle \bv^{(k)}, \bv_*\rangle$ and similarly for $\theta_{\bu_k}$. This suggests that, under some conditions, the sequence of iterates can contract to the true solution ($\theta = 0$): specifically, in the noiseless ($\Ecal = 0$) case, we recall that
\[2|1 - \cos \theta| \leq |\sin \theta| \quad \text{ for all } |\theta| \in [0, 2\tan^{-1}(0.5) \approx [0^{\circ}, 53.1^{\circ}]\]
In low-dimensions, this condition can typically be obtained by random initialization; in moderate- to high-dimensional settings, a more nuanced initialization is required. Under this ``shrinkage'' condition, the proof follows by repeated substitution of Equations (\ref{eqn:sin_v}-\ref{eqn:sin_u}) into each other. 

The extension to the noisy case is more difficult. It does not suffice for the passage between $\sin(\cdot)$ and $\cos(\cdot)$ to be non-contractive, because the effect of the noise term $\opnorm{\Ecal}$ can accumulate over repeated iterations. To handle the noisy case, we require a strict contraction at each step. Specifically, we assume that:
\begin{equation}
    \begin{aligned}
    2|1 - \cos \theta_{\bu_k}| \leq c_u |\sin \theta_{\bu_k}| \quad \text{ for all $k$, for some $c_u < 1$} \\ 
    2|1 - \cos^4 \theta_{\bv_k}| \leq c_v |\sin \theta_{\bv_k}| \quad \text{ for all $k$, for some $c_v < 1$.}
    \end{aligned}
     \label{eqn:contractions}
\end{equation}
Then we have the following bounds for the $\bv$ iterates: 
\begin{align}
    |\sin \theta_{\bv_{k+1}}| &\leq 2|1 - \cos \theta_{\bu_k}| + \frac{2\opnorm{\Ecal}}{d} \notag\\ 
    &\leq c_u|\sin \theta_{\bu_k}| + \frac{2\opnorm{\Ecal}}{d} \notag\\ 
    &\leq c_u\left(2|1 - \cos^4\theta_{\bv_{k}}| + \frac{4\opnorm{\Ecal}}{d} + \frac{2\opnorm{\Ecal}^2}{d^2}\right) + \frac{2\opnorm{\Ecal}}{d} \notag\\
    &\leq c_uc_v |\sin \theta_{\bv_{k}}| + \frac{4c_u\opnorm{\Ecal}}{d} + \frac{2c_u\opnorm{\Ecal}^2}{d^2} + \frac{2\opnorm{\Ecal}}{d} \label{eqn:v_bound}
\end{align}
Iterating this, we find that: 
\[|\sin \theta_{\bv_k}| \leq (c_u c_v)^k |\sin \theta_{\bv_1}| + \left(\frac{4c_u\opnorm{\Ecal}}{d} + \frac{2c_u\opnorm{\Ecal}^2}{d^2} + \frac{2\opnorm{\Ecal}}{d}\right)\sum_{i=0}^{k-1} (c_uc_v)^i\]
We highlight here that the convergence of the non-stochastic part of the error is geometric in $c_uc_v$. Letting $k \to \infty$, the first term vanishes and the second term can be simplified as
\begin{align*}
    |\sin \theta_{\bv_k}| &\leq \frac{1}{1 - c_uc_v}\left(\frac{4c_u\opnorm{\Ecal}}{d} + \frac{2c_u\opnorm{\Ecal}^2}{d^2} + \frac{2\opnorm{\Ecal}}{d}\right) \\
    &\leq \frac{6\frac{\opnorm{\Ecal}}{d} + 2\frac{\opnorm{\Ecal}^2}{d^2}}{1 - c_uc_v} \\
    &\leq \frac{6\frac{\opnorm{\Ecal}}{d} + 2\frac{\opnorm{\Ecal}}{d}}{1 - c_uc_v} \\
    &= \frac{8\opnorm{\Ecal}}{d(1 - c_u c_v)}
\end{align*}
when $\frac{\opnorm{\Ecal}}{d} < 1$. Interestingly, the explicit term measuring the quality of the initialization ($\sin \theta_{\bv_0}$) vanishes, but the quality of the initialization persists implicitly in the $c_u$ and $c_v$ terms. 

Bounds for the $\bu$ iterates can be attained similarly: 
\begin{align*}
    |\sin \theta_{\bu_k}| &\leq 2|1 - \cos^4 \theta_{\bv_{k}}| + \frac{4\opnorm{\Ecal}}{d} + \frac{2\opnorm{\Ecal}^2}{d^2} \\
    &\leq c_v |\sin \theta_{\bv_{k}}| + \frac{4\opnorm{\Ecal}}{d} + \frac{2\opnorm{\Ecal}^2}{d^2} \\
    &\leq c_v \left(2|1 - \cos \theta_{\bu_{k-1}}| + \frac{2\opnorm{\Ecal}}{d}\right) + \frac{4\opnorm{\Ecal}}{d} + \frac{2\opnorm{\Ecal}^2}{d^2} \\
    &\leq c_u c_v|\sin \theta_{\bu_{k-1}}| + \frac{2c_v\opnorm{\Ecal}}{d} + \frac{4\opnorm{\Ecal}}{d} + \frac{2\opnorm{\Ecal}^2}{d^2}
\end{align*}
Iterating, this yields: 
\[|\sin \theta_{\bu_k}| \leq (c_uc_v)^k |\sin \theta_{\bu^{(0)}}| + \left(\frac{2c_v\opnorm{\Ecal}}{d} + \frac{4\opnorm{\Ecal}}{d} + \frac{2\opnorm{\Ecal}^2}{d^2}\right)\sum_{i=0}^{k-1}(c_uc_v)^i\]
As before, this implies convergence of the non-stochastic error at a rate geometric in $c_uc_v$. In the $k \to \infty$ limit, this becomes: 
\begin{align*}
    |\sin \theta_{\bu_k}| &\leq \frac{1}{1-c_uc_v}\left(\frac{2c_v\opnorm{\Ecal}}{d} + \frac{4\opnorm{\Ecal}}{d} + \frac{2\opnorm{\Ecal}^2}{d^2}\right) \\
     &\leq \frac{6\frac{\opnorm{\Ecal}}{d} + 2\frac{\opnorm{\Ecal}^2}{d^2}}{1 - c_uc_v} \\
    &\leq \frac{6\frac{\opnorm{\Ecal}}{d} + 2\frac{\opnorm{\Ecal}}{d}}{1 - c_uc_v} \\
    &= \frac{8\opnorm{\Ecal}}{d(1 - c_u c_v)}
\end{align*}
matching the $\theta_{\bv_k}$ bounds given above. 

The analysis in this section extends naturally to the general rank-$r$ case: algebraically, it is simplier to work with the root mean Frobenius error $\|\sin \Theta(\bV_*, \bV^{(k+1)})\|_F / \sqrt{r}$ rather than the total Frobenius error to eliminate the extra factor of $\sqrt{r}$. Under this scaling, the right hand sides of each step match those given above. The resulting error bound is then: 
\[\|\sin \Theta(\bv_*, \bv^{(k)})\|_F \leq \frac{8\sqrt{r}\ropnorm{\Ecal}}{d(1 - c_u c_v)} \text{ as } k \to \infty\]

Finally, to get the MSE bounds given in the statement of this theorem, we recall that
\[\sin \angle(\bu_*, \hat{\bu}) = c \Longleftrightarrow \min_{\epsilon = \pm 1} \frac{\|\epsilon \bu^* - \hat{\bu}\|_2}{\sqrt{2}} = c\]
Hence, 
\begin{align*}
   \min_{\epsilon = \pm 1} \|\epsilon \bu^* - \hat{\bu}\|_2 &\leq \frac{8\sqrt{2}\opnorm{\Ecal}}{d(1 - c_uc_v)} \\
   \min_{\epsilon = \pm 1} \|\epsilon \bv^* - \hat{\bv}\|_2 &\leq \frac{8\sqrt{2}\opnorm{\Ecal}}{d(1 - c_uc_v)}
\end{align*}
Or in the rank-$r$ case: there exists an orthogonal matrix $\bO$ such that: 
\[\|\bV^*\bO - \hat{\bV}\|_F \leq \frac{8\sqrt{2r}\ropnorm{\Ecal}}{d(1 - c_uc_v)}\]
The passage from $\sin(\cdot)$ bounds to $\ell_2$ error is standard: see, \emph{e.g.}, the discussion of Equation A5 in the paper by \citet{Yu:2015}. 

\noindent \textbf{IV. Condition Checking.} In the previous section, we assumed in Equation \eqref{eqn:contractions} that we had a ``shrinkage factor'' of at least $\max(c_u, c_v) < 1$ at each iteration. This assumption may appear strenuous but it follows naturally from the assumption on the initialization $\bu$ in the statement of our theorem.  Specifically, we can combine an initialization condition with the contraction condition of Equation \eqref{eqn:v_bound} to ensure that we stay in the ``contraction interval'' given by $|\sin \theta| \in [0, 2\tan^{-1}(0.5))$. Specifically, we note that at iteration $k$, we have
\begin{align*}
    |\sin \theta_{\bv_k}| &= (c_uc_v)^k |\sin \theta_{\bv_1}| + \left(\frac{2c_v\opnorm{\Ecal}}{d} + \frac{4\opnorm{\Ecal}}{d} + \frac{2\opnorm{\Ecal}^2}{d^2}\right)\frac{1 - (c_uc_v)^k}{1 - c_uc_v} \\ &\leq |\sin \theta_{\bv_1}| + \frac{8\opnorm{\Ecal}}{d(1 - c_uc_v)} \quad \text{ for all } k = 1, 2, 3, \dots
\end{align*}
Hence, if we want to remain in the contraction interval for all $k$, it suffices to have 
\[|\sin \theta_{\bv_1}| + \frac{8\opnorm{\Ecal}}{d(1 - c_uc_v)} \leq 2\tan^{-1}(0.5)\]
or, equivalently, 
\[|\sin \theta_{\bv_1}| \leq 2\tan^{-1}(0.5) - \frac{8\opnorm{\Ecal}}{d(1 - c_uc_v)}.\]
To connect this to our initialization condition on $\bu^{(0)}$, we use Equation \eqref{eqn:sin_v} one final time to find that it suffices to have
\[2|1 - \cos \theta_{\bu^{(0)}}| + \frac{2\opnorm{\Ecal}}{d} \leq 2\tan^{-1}(0.5) - \frac{8\opnorm{\Ecal}}{d(1 - c_uc_v)}.\]
Rearranging this, we obtain the initialization condition on $\bu$:
\[|1 - \cos \theta_{\bu^{(0)}}| \leq \tan^{-1}(0.5) - \frac{5\opnorm{\Ecal}}{d(1 - c_uc_v)},\]
letting $c = c_uc_v$ be the arbitrary constant in the statement of the proposition.

The general rank $r$-case follows immediately. \qedhere
\end{proof}

Next, we prepare to transfer from the ``deterministic'' Proposition \ref{appprop:consistency} to the more familiar stochastic setting of Theorems \ref{thm:consistency} and \ref{thm:vconsistency} via a standard concentration bound on $\ropnorm{\Ecal}$. Our key tool is Proposition \ref{prop:ropnorm}, which we restate and prove with more detail here for convenience: 

\propropnorm*

\begin{proof} For the deterministic bound, \[\ropnorm{\Xcal} \leq k \sqrt{T}\max_i \left|\lambda_{\max}(\Xcal_{\cdot\cdot i})\right|,\] we note that

\begin{align*}
\ropnorm{\Xcal} &= \max_{\bu, \bV \in \overline{\B}^T \times \Vcal^{p \times r}} \left|\left\langle [\Xcal; \bV], \bu\right\rangle\right| \\
    &= \max_{\bu \in \overline{\B}^T, \bV \in \Vcal^{p \times r}} \left|\left\langle \Tr(\bV^T\Xcal_{\cdot \cdot i}\bV)_i, \bu \right\rangle \right|\\
    &\leq \max_{\bu \in \overline{\B}^T \bV_i \in \Vcal^{p \times r}}\left|\left\langle \Tr(\bV_i^T\Xcal_{\cdot \cdot i}\bV_i)_i, \bu \right\rangle \right|\\
    &= \max_{\bu \in \overline{\B}^T \bV_i \in \Vcal^{p \times r}}\left|\left\langle \left(\sum \lambda_{\max:r}(\Xcal_{\cdot\cdot i})\right)_i, \bu \right\rangle \right|\\
    &\leq r \max_{\bu \in \overline{\B}^T} \left|\langle \left(\lambda_{\max}(\Xcal_{\cdot\cdot i})\right)_i, \bu \rangle \right|
\end{align*}
where the first inequality follows by allowing different $\bV_i$ for each slice of $\Xcal$, which can be independently set to the leading $r$-eigenvectors of $\Xcal_{\cdot\cdot i}$, so that the trace is then given by the sum of the top $r$ eigenvalues of the $i^{\text{th}}$ slice. The second inequality follows by bounding the sum by $r$ times the maximum eigenvalue. The maximum here is obtained by $\bu \propto (\lambda_{\max}(\Xcal_{\cdot\cdot i}))_i$ and gives the value $\|(\lambda_{\max}(\Xcal_{\cdot\cdot i}))_i\|_2$. Because the elements of this random vector are not mean-zero, this quantity is slightly cumbersome to bound. An acceptable bound can be obtained by recalling that $\|\bx\|_2 \leq \sqrt{T} \|\bx\|_{\infty}$ for any $\bx \in \R^T$ and then bounding the maximum element. Specifically, we have
\begin{align*}
    \ropnorm{\Xcal} &\leq r \|(\lambda_{\max}(\Xcal_{\cdot\cdot i}))_i\|_2 \\
                    &\leq r\sqrt{T} \|(\lambda_{\max}(\Xcal_{\cdot\cdot i}))_i\|_{\infty} \\
                    &= r\sqrt{T} \max_i \lambda_{\max}(\Xcal_{\cdot\cdot i})
\end{align*}

Next, to get the stochastic claim, we need to control $\lambda_{\max}(\Xcal_{\cdot \cdot i})$ for each $i$ and then apply a union bound to control $\max_i \lambda_{\max}(\Xcal_{\cdot \cdot i})$. This follows straightforwardly from well-known bounds on the operator norm of sub-Gaussian random matrices.  Specifically, note that each slice of $\Xcal$ ($\Xcal_{\cdot\cdot i}$) is a symmetric matrix with $\sigma^2$-sub-Gaussian entries. Standard results allow us to characterize the operator norm quite accurately: specifically, we have, for fixed $i$,
\begin{align*}
    \opnorm{\Xcal_{\cdot \cdot i}} \lesssim \sigma\left(\sqrt{p} + u\right)
\end{align*}
with probability at least $1 - 4\exp(-u^2)$: see, \emph{e.g.}, Corollary 4.4.8 in the text by \citet{Vershynin:2018}. Combining this with the eigenvalue bound above,  we have
\begin{align*}
    P\left\{\ropnorm{\Xcal} \gtrsim \sigma r\sqrt{T}\left(\sqrt{p} + u\right)\right\} &\leq T P\left\{\opnorm{\Xcal_{\cdot \cdot 1}} \gtrsim \sigma \left(\sqrt{p} + u\right)\right\} 
\end{align*}
by a standard union bound (Boole's inequality) applied to the $T$ independent slices of $\Xcal$.\footnote{Let $A_i$ be the event $\opnorm{\Xcal_{\cdot \cdot i}} \gtrsim r\sigma(\sqrt{p} + u)$. Then $A = \bigcup_i A_i$ is the event $\ropnorm{\Xcal} \gtrsim r\sigma(\sqrt{p} + u)$ and Boole's inequality gives us: 
\[P(A) \leq \sum_i P(A_i) = T P(A_1)\] with the last inequality following from independence.} Standard manipulations then yield:
\begin{align*}
    P\left\{\opnorm{\Xcal_{\cdot \cdot 1}} \lesssim \sigma \left(\sqrt{p} + u\right)\right\} &\geq 1 - 4e^{-u^2} \\
    1 - P\left\{\opnorm{\Xcal_{\cdot \cdot 1}} \lesssim \sigma \left(\sqrt{p} + u\right)\right\} &< 1 - \left[1 - 4e^{-u^2}\right] \\
    P\left\{\opnorm{\Xcal_{\cdot \cdot 1}} \gtrsim \sigma \left(\sqrt{p} + u\right)\right\} &< 4e^{-u^2}
\end{align*}
Hence, 
\[P\left\{\ropnorm{\Xcal} \gtrsim \sigma r\sqrt{T} \left(\sqrt{p} + u\right)\right\} < 4Te^{-u^2}\]
or equivalently
\[P\left\{\ropnorm{\Xcal} \lesssim \sigma r\sqrt{T}\left(\sqrt{p} + u\right)\right\} \geq 1 - 4e^{-u^2 + \log T}.\]
Letting $u = \sqrt{\delta^2 + \log T} \Leftrightarrow -u^2 + \log T = -\delta^2$, we obtain the bound 
\[P\left\{\ropnorm{\Ecal} \lesssim \sigma r\sqrt{T}\left(\sqrt{p} + \sqrt{\delta^2 + \log T}\right)\right\} \geq 1 - 4e^{-\delta^2}.\]
Finally, recalling that $\sqrt{a^2 + b} \leq a + \sqrt{b}$ for all $a, b > 1$, we have 
\[P\left\{\ropnorm{\Ecal} \lesssim \sigma r\sqrt{T}\left(\sqrt{p} +\sqrt{\log T} + \delta\right)\right\} 
\geq 1 - 4e^{-\delta^2}. \qedhere\]
\end{proof} 

With these two results, we are now ready to prove Theorems \ref{thm:consistency} and \ref{thm:vconsistency} which we restate here for convenience of the reader: 

\consistency*

\vconsistency*

\begin{proof} Under the stated conditions, Proposition \ref{appprop:consistency} implies 
\begin{align*}
  \min_{\epsilon = \pm 1} \|\epsilon \bu^* - \hat{\bu}\|_2 &\leq \frac{8\sqrt{2}\ropnorm{\Ecal}}{d(1 - c)} \\
  \|\bV^*\bO - \hat{\bV}\|_F &\leq \frac{8\sqrt{2r}\ropnorm{\Ecal}}{d(1 - c)}
\end{align*}
so long as $\ropnorm{\Ecal} < d$ and 
\[|1 - \langle \bu^{(0)}, \bu_*\rangle | \leq \tan^{-1}(0.5) - \frac{5 \ropnorm{\Ecal}}{d(1 - c)}\] for some arbitrary $c < 1$. Furthermore, under the same conditions, Proposition \ref{prop:ropnorm} applied to $\Ecal$ implies that $\ropnorm{\Ecal} \lesssim \sigma r\sqrt{T}(\sqrt{p} + \sqrt{\log T} + \delta)$ with probability at least $1 - 4e^{-\delta^2}$. Substituting this into the error bounds we have: 
\begin{align*}
  \min_{\epsilon = \pm 1} \|\epsilon \bu^* - \hat{\bu}\|_2 &\lesssim \frac{\sigma r\sqrt{T}(\sqrt{p} + \sqrt{\log T} + \delta)}{d(1 - c)} \\
  \|\bV^*\bO - \hat{\bV}\|_F &\lesssim \frac{\sqrt{r}\sigma r\sqrt{T}(\sqrt{p} + \sqrt{\log T} + \delta)}{d(1 - c)}.
\end{align*}
If we normalize the right hand side of each quantity to ``root mean squared error'' instead of total $\ell_2$ error, we have: 
\begin{align*}
  \min_{\epsilon = \pm 1} \|\epsilon \bu^* - \hat{\bu}\|_2/\sqrt{T} &\lesssim \frac{\sigma r \sqrt{T}
  (\sqrt{p} + \sqrt{\log T} + \delta)}{d\sqrt{T}(1 - c)} \\ &\lessapprox \frac{\sigma \sqrt{p}}{d(1 - c)}\\
  \|\bV^*\bO - \hat{\bV}\|_F / \sqrt{pr} &\lesssim \frac{\sigma r\sqrt{T}(1 + \sqrt{(\log T)/pr} + \delta/\sqrt{pr})}{d(1 - c)} \\ &\lessapprox \frac{\sigma \sqrt{T}}{d(1 - c)}
\end{align*}
with bounds holding with high probability. \qedhere
\end{proof}

\section{Additional Material for Section \ref{sec:empirical} - \nameref{sec:empirical}}\label{app:empirical}


\subsection{Additional Case Study: Correlation of International Stock Markets} \label{app:stocks}
\begin{figure}[ht]
\centering
\includegraphics[width=\textwidth]{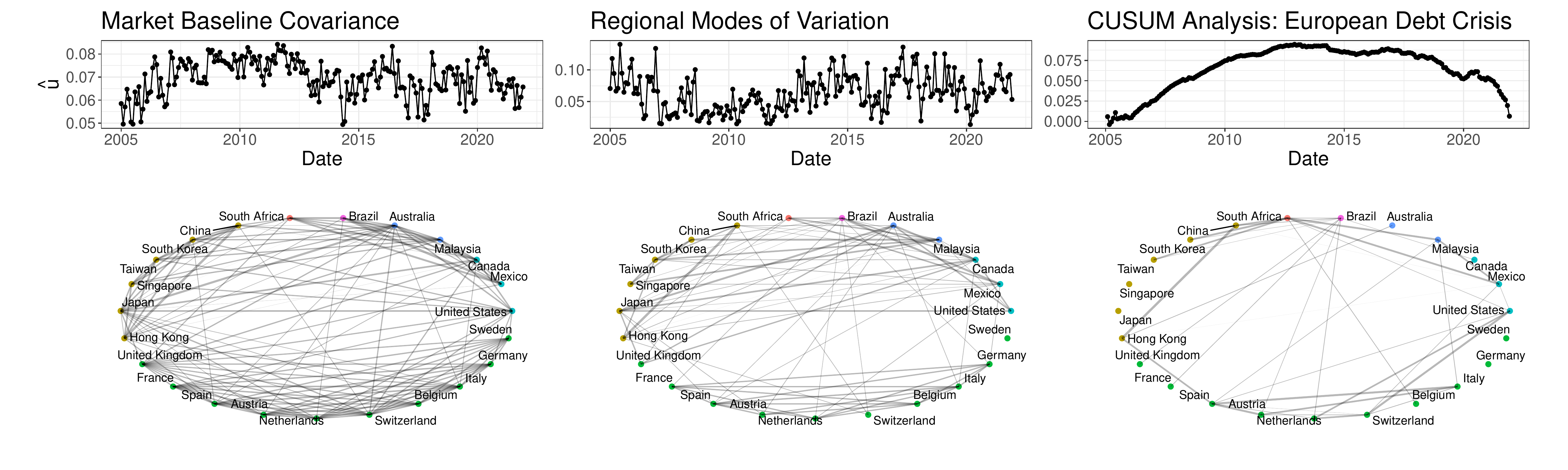}
\caption{Application of SS-TPCA to the stock market covariance example described in Subsection \ref{app:stocks}. Baseline market behavior (right column) reflects world-wide economic trends applying to all indices roughly equally; the most significant differences in behavior (center column) capture economic changes in different regions of the world, while the most significant change over time (right column) highlights the divergence of Eurozone stock indices during the European Debt Crisis (major period roughly 2009-2015, though ongoing until 2020). Top row is the estimated time-loading vector $\hat{\bu}$ which is roughly constant for the global economy factor; bottom row visualizes the estimated networks $\hat{\bV} \circ \hat{\bV}$ with small edges thresholded for clarity. Rank-$r$ selected via optimal BIC for each model.}
\label{fig:stocks}
\end{figure}

We apply SS-TPCA proposals to analyze the behavior of 23 international stock markets, for the period from January 2005 to November 2021. For each month, we construct a $23\times 23$ network with edges given by the correlation among (US Dollar denominated) total returns of national stock market indices. Repeating this process over our time period gives a semi-symmetric tensor of dimension $23 \times 23 \times 204$, each slice of which is positive definite by construction. We then apply SS-TPCA to identify baseline patterns of market behavior, significant deviations from baseline, and to examine evidence for changes over our 16 year period of interest. Our results are presented in Figure \ref{fig:stocks}.

Not surprisingly, the baseline behavior identified by applying SS-TPCA to the raw data tensor identifies a baseline network with nearly constant edge strength. This baseline captures the well-known effect of overall market volatility not being isolated in a single market segment or country: this effect is essentially the same as the market or ``beta'' terms from the Capital Asset Pricing Model and its variants \citep{Fama:2004}, 
though here calculated from SS-TPCA rather than a cross-sectional regression model as is standard in the econometric literature. Examining the associated time loading vector ($\hat{\bu}$), we see that the importance of this factor is essentially unchanged throughout our time period. 

Applying SS-TPCA to the residuals of our baseline analysis, we next identify the major patterns that drive deviations from the market-wide factor. As Figure \ref{fig:stocks} shows, this analysis identifies major geographic regions, implying that, after global effects, the most important market patterns correspond to well-known regional effects (Europe, Asia, and North America). Examining the associated time loading vector, we see that the importance of these regional differences was significantly reduced in the period from 2007 to 2010, roughly capturing the period of the global financial crisis and the systemic shocks arising in that period. 

Finally, we perform a \cusum analysis on our data to investigate the possibility of market structure changes over our period of interest. While the principal network estimated in this analysis ($\hat{\bV} \circ \hat{\bV}$) is less immediately interpretable than those from the previous two analyses, we note that it is primarily concentrated on edges connecting southern European nations, such Spain and Italy, at the heart of the post-2010 Eurozone debt crisis, with a secondary community centered around China and other rapidly industrializing nations. The \cusum vector ($\hat{\bu}$) shows that the effects of these changes were most pronounced in the period from 2011 to 2016, consistent with the height of Eurozone debt challenges. As we would expect, these effects dissipate rapidly in early 2020 as stock markets began to react to the effects of the COVID-19 pandemic. 

\section{Additional Comments on SS-TPCA}
\subsection{Stable Initialization}
As discussed in Section \ref{sec:theory}, the most important assumption of Theorems \ref{thm:consistency} and \ref{thm:vconsistency} is the initialization condition on $\bu^{(0)}$; this inequality is quite difficult to check empirically as it involves the unknown parameter $\bu_*$, the norm of the unobserved noise tensor $\Ecal$, and an arbitrary constant $c < 1$. As mentioned there, the initialization condition on $\bu^{(0)}$ is more easily satisfied for SS-TPCA applied to network-valued time series than might otherwise be expected, as networks are generally highly auto-correlated over time, implying that $\bu_*$ lies near the constant vector $\bu_T / \sqrt{T}$. The following definition formalizes this intuition, giving an alternative statement of the consistency conditions: 
\begin{appendixproposition} \label{prop:stable_init} Suppose $\bu_*$ is a $\delta$-\emph{stable} unit vector in $\R^T$, in the sense that
\[|\langle \bu_*, \bone_T / \sqrt{T} \rangle - 1| \leq \delta,\] where $\bone_T$ is the all-ones vector in $\R^T$ for \[\delta \leq \tan^{-1}(0.5) - \frac{5 \sigma \sqrt{p}}{d(1 - c)}\] Then, the initialization condition of Theorems \ref{thm:consistency} and \ref{thm:vconsistency} holds with high probability when Algorithm \ref{alg:single_factor_sstpca} is initialized at $\bu_0 = \bone_T / \sqrt{T}$.
\end{appendixproposition} 
\noindent This \emph{stability condition} implies that $\bu_*$ is ``approximately constant'' and that the (scaled) all-ones vector provides a good initialization to Algorithm \ref{alg:single_factor_sstpca}. We expect this assumption to be a reasonable one in most applications: \emph{e.g.}, if $\Xcal$ represents correlation of financial assets or linkages in a social network over multiple time periods, we expect the main signal to remain roughly constant over time. It would be highly unusual for correlated financial instruments to suddenly become anti-correlated or for all connections in a social network to vanish and be replaced at random. While the stability condition may be expected to hold in many scenarios and to give guidance for initializing Algorithm \ref{alg:single_factor_sstpca}, we have found that random initialization also performs quite well in practice, as shown by the numerical results of Section \ref{sec:empirical}.

By analogy with the stability condition, we refer to the choice of $\bu_0 = \bone_T / \sqrt{T}$ as ``stable initialization'' for Algorithm \ref{alg:single_factor_sstpca}. Stable initialization admits a particularly nice interpretation of the first $\bV$-update in Algorithm \ref{alg:multi_factor_sstpca} which we highlight below: 
\begin{appendixremark} \label{rem:stable_init}
  Under the stable initialization condition $\bu_0 = \bone_T / \sqrt{T}$ and the low-rank + noise model assumed in Theorems \ref{thm:consistency} and \ref{thm:vconsistency}, the first step of Algorithm \ref{alg:single_factor_sstpca} implicitly estimates $\bV_* \circ \bV_*$ \emph{via} the sample average of $T$ observations; specifically, 
  \[d \bV_* \circ \bV_* \approx d\bV_* \circ \bV_* \langle \bu_0, \bu_* \rangle + \Ecal \ttv{3} \bu_0\]
  For large values of $\langle \bu_0, \bu_* \rangle$, the first term dominates the second, implying an effective initialization. Specifically, the stable-initialization condition implies \[\|\sin \Theta(\bV^*, \bV^{(1)})\|_F \leq \frac{2r^{1/2}\opnorm{\tilde{\bE}}}{d\langle \bu_*, \bu_0\rangle}\]
  where $\tilde{\bE} = \Ecal \ttv{3} \bu_0$ is a symmetric matrix with independently $\sigma^2$-sub-Gaussian elements, and hence $\opnorm{\tilde{\bE}}$ is of the order $\sigma \sqrt{p}$ with high probability. 
\end{appendixremark}
\noindent Of course, the variance of this initial estimator could be improved by knowing the ``oracle'' initialization $\bu_0 = \bu_*$, but even this simple initialization scheme works well in practice. This heuristic analysis can also be extended to the case of random initialization by noting that $\langle \bu_*, \bu_0 \rangle$ is unlikely to be exactly zero by chance and so some signal is likely to persist in the initial $\bV$-update by chance. When $d$ is sufficiently large, this residual signal can be expected to dwarf the noise term and Algorithm \ref{alg:single_factor_sstpca} will again be well initialized.

\subsection{Rank Selection}
As with matrix PCA, selection of the optimal ranks $(r_1, \dots, r_K)$ for use in Algorithm \ref{alg:multi_factor_sstpca} is a challenging problem for which a fully satisfactory answer is not readily available. Many heuristics have been proposed for rank selection in matrix PCA \citep{Cattell:1966,Owen:2009,Josse:2012,Choi:2017}, typically based on decreasing explanatory power or a more general information criterion-based model selection framework. In our experiments, we have found that a greedy approach where each rank $r_i$ is chosen by maximizing the BIC performs well \citep{Bai:2018}, but tensor analogues of data-driven approaches for PCA have also been shown to perform well in similar settings \citep{Sedighin:2021}. Unlike matrix PCA, tensor decomposition methods are not based on ordered singular vectors and cannot guarantee nestedness or orderedness of the estimated factors \emph{a priori}. In practice, we have found that the greedy structure of Algorithm \ref{alg:multi_factor_sstpca}, which combines a power method and a strict deflation step, produces reasonably well-ordered factors (decreasing values of $\{d_i\}$) and gives nestedness by construction. Similar behavior has been observed in other power method contexts, including differentially private PCA \cite{Hardt:2014}, sparse matrix PCA \citep{Journee:2010}, and regularized tensor factorization \citep{Allen:2013, Allen:2012}. Because Algorithm \ref{alg:multi_factor_sstpca} does not enforce orthogonality of the estimated components $\bV_i$ and $\bV_j$ as in matrix PCA, the specific choice of rank is less important than in matrix PCA, so long as the rank is as high as the rank of the target network, as has been recently noted for several other decomposition schemes, \emph{e.g.}, non-negative matrix factorization \citep{Kawakami:2021}. Note in particular that if $\bV_*$ is of rank $r$ but we have iterates $\bV^{(k)}$ of rank $r' > r$, the inner product $\langle \bV_*, \bV^{(k)} \rangle$ implicit in the $\bu$-update (see part II of the proof of Proposition \ref{appprop:consistency}) will, in expectation, zero out the additional degrees of freedom. The flexible eigenvalue scheme alluded to in Section \ref{sec:sstpca}, where $\bV$ is allowed to take values in the interior of the Stiefel manifold, is particularly useful in this case. 


\subsection{Robustness to a Dynamic Adversary} 
The proof of Proposition \ref{appprop:consistency} implies that SS-TPCA is actually consistent under a more sophisticated noise model than discussed in the main text of our paper. In particular, we have the following:
\begin{appendixproposition} \label{appprop:consistency_adversarial}
With $\Xcal_*$ generated as in Proposition \ref{appprop:consistency}, suppose the single-factor SS-TPCA algorithm is run with the following noisy updates: 
\begin{align*}
    \bV^{(k+1)} &= r\textsf{-eigen}\left(\Xcal_* \ttv{3} \bu^{(k)} + \bE_V^{(k)}\right) \\
    \bu^{(k+1)} &= \textsf{Norm}\left([\Xcal_*; \bV^{(k+1)}] + \be_u^{(k+1)}\right) 
\end{align*}
where $\bE_V^{(k)}, \be_u^{(k)}$ may differ at each iteration and may be chosen adversarially. So long as 
\[\max\left\{\max_k \opnorm{\bE_V^{(k)}}, \max_k \|\be_u^{(k)}\|_2\right\} \leq 1.1dr\]
and the initialization condition is satisfied, the same accuracy bounds hold.
\end{appendixproposition}
\citet{Hardt:2014} noted a similar property in their analysis of the noisy power method. While this type of \textbf{dynamic adversarial robustness} is not yet well explored in the literature, it is important for the network science applications motivating this work and we believe it to be worthy of additional study. 

\printbibliography[title={Additional References}]
\end{refsection}
\end{document}